\documentclass[0.5pt,letterpaper]{article}
\pdfoutput=1
\usepackage{fullpage}
\usepackage{geometry}

\usepackage[numbers]{natbib}

\usepackage{microtype}
\usepackage{graphicx}
\usepackage{booktabs} %

\usepackage{hyperref}
\usepackage{bm}

\usepackage[utf8]{inputenc} %
\usepackage[T1]{fontenc}    %
\usepackage{xcolor}

\definecolor{mydarkblue}{rgb}{0,0.08,0.45}

\usepackage{array}
\usepackage{url}            %
\usepackage{amsfonts}       %
\usepackage{nicefrac}       %
\usepackage{microtype}      %

\usepackage{amsmath}
\usepackage{multirow}
\usepackage{color}
\usepackage{bm}
\usepackage{amsthm}
\usepackage{mathtools}
\usepackage{multicol}
\usepackage{stmaryrd}
\usepackage{grffile}%
\usepackage[scriptsize,hang]{subfigure}
\def\rot#1{\rotatebox[origin=c]{90}{#1}}
\newcommand{\punt}[1]{}

\newtheorem{theorem}{Theorem}

\def\argmin{\mathop{\rm arg\,min}}

\newcommand{\reals}{\mathbb{R}}

\newcommand{\expect}{\mathbb{E}}

\newcommand{\kl}{\textrm{KL}}

\def\argmin{\mathop{\rm arg\,min}}

\newcommand{\bq}{\begin{equation}}
\newcommand{\eq}{\end{equation}}
\newcommand{\ba}{\begin{eqnarray}}
\newcommand{\ea}{\end{eqnarray}}

\def\R{{\reals}}

\newcommand{\mcal}[1]{\mathcal{#1}}
\newcommand{\remove}[1]{}

\newcommand{\ie}{\textit{i}.\textit{e}.}
\newcommand{\eg}{\textit{e}.\textit{g}.}
\newcommand{\etc}{\textit{etc}}

\newcommand{\cmark}{\ding{51}}%
\newcommand{\xmark}{\ding{55}}%

\usepackage{adjustbox}
\usepackage{array}

\newcolumntype{R}[2]{%
    >{\adjustbox{angle=#1,lap=\width-(#2)}\bgroup}%
    l%
    <{\egroup}%
}
\renewcommand*\rot[2]{\multicolumn{1}{R{#1}{#2}}}%

\usepackage{algorithm}
\usepackage[noend]{algpseudocode}
\usepackage{wrapfig}
\usepackage{siunitx} %
\usepackage{pifont}
\usepackage{enumitem} %
\usepackage{bm}
\usepackage{url}

\definecolor{mydarkblue}{rgb}{0,0.08,0.45}
\hypersetup{ %
    colorlinks=true,
    linkcolor=mydarkblue,
    citecolor=mydarkblue,
    filecolor=mydarkblue,
    urlcolor=mydarkblue}

\usepackage{tabularx}
\makeatletter
\newcommand{\multiline}[1]{%
  \begin{tabularx}{\dimexpr\linewidth-\ALG@thistlm}[t]{@{}X@{}}
    #1
  \end{tabularx}
}
\makeatother

\begin{document}

\title{Constrained Instance and Class Reweighting for Robust Learning under Label Noise}

\author{Abhishek Kumar\\
{\tt\normalsize abhishk@google.com}
\and
Ehsan Amid\\
{\tt\normalsize eamid@google.com}
}

\date{Google Research, Brain Team}

\maketitle 
\begin{abstract}
Deep neural networks have shown impressive performance in supervised learning, enabled by their ability to fit well to the provided training data. However, their performance is largely dependent on the quality of the training data and often degrades in the presence of noise. We propose a principled approach for tackling label noise with the aim of assigning importance weights to individual instances and class labels. Our method works by formulating a class of constrained optimization problems that yield simple closed form updates for these importance weights. The proposed optimization problems are solved per mini-batch which obviates the need of storing and updating the weights over the full dataset. Our optimization framework also provides a theoretical perspective on existing label smoothing heuristics for addressing label noise (such as label bootstrapping). We evaluate our method on several benchmark datasets and observe considerable performance gains in the presence of label noise.
\end{abstract}

\vspace{-4mm}
\section{Introduction}
\vspace{-2mm}
\label{sec:intro}
Deep neural networks have been quite successful in driving impressive performance gains in several real-world applications. However, %
overparametrized deep networks can easily overfit to noisy or corrupted labels \citep{zhang2016understanding},
and training with noisy labels often leads to degradation in generalization performance on clean test data. Unfortunately, noisy labels can naturally appear in several real world scenarios, such as labels obtained from the internet, noisy human annotations, automatic labels obtained from legacy rule based systems or from machine learned systems trained on obsolete or shifted data distributions, \etc. %
The significance of this problem has inspired a long line of work on robust learning under label noise. Past attempts at addressing this problem include identifying and correcting the label noise \citep{hendrycks2018using,extendedT}, designing robust loss functions \citep{bitempered,peerloss,ma2020normalized}, label smoothing~\citep{szegedy2016rethinking,lukasik2020does}, regularization~\citep{azadi2015auxiliary,miyato2018virtual,foret2021sharpnessaware}, reweighting the training examples \citep{ren2018learning,bar2021multiplicative,majidi2021exponentiated}, curriculum learning \citep{jiang2018mentornet,saxena2019data}, \etc. 

In this work, we propose a novel and principled method for dynamically assigning importance weights to each instance and class label in a minibatch. 
We propose a class of constrained optimization problems where we control for the deviation of these importance weights from a reference distribution (\eg, uniform) as measured by a divergence of choice. We obtain simple closed form updates for the importance weights for several common divergence measures such as $\alpha$-divergences \citep{chernoff-alpha,Cichocki_2010}. %
We also propose a novel method of using these importance weights with Mixup \citep{zhang2017mixup} that further yields significant empirical improvements. 
We evaluate the proposed method on standard benchmarks used in earlier works, %
comparing with several state-of-the-art approaches for learning with label noise, and observe considerable improvements in the metric of interest (test accuracy on clean data). %
Our contributions are summarized as follows:
\begin{itemize}[noitemsep,topsep=0pt,leftmargin=6mm]
  \item We propose a principled optimization based formulation for instance and class reweighting for label noise that results in simple closed form updates.
  \item The proposed method does not maintain the weights across the whole training set and hence fits easily in standard training pipelines with little computational or memory overhead. This is an added benefit compared to some recently proposed methods that keep a record of importance weights over the complete training set \citep{saxena2019data,bar2021multiplicative,majidi2021exponentiated,liu2020early} which results in increased overhead, particularly in large scale production settings. These methods are also not applicable in streaming settings where each sample is visited only once. 
  We show a comparison of several recent methods %
  in Table \ref{tab:compare}. 
  \item We show that some earlier heuristics for label noise, such as label bootstrapping \cite{reed2014training,arazo2019unsupervised} can be naturally derived from our optimization framework for specific choice of divergence.
  \item Empirically, we observe that our method provides significant improvements in different noise settings, more so over methods that do not require extra storage or training an extra network. Our method can also be combined with other complementary approaches such as Mixup to further their benefits. 
\end{itemize}

\vspace{-1mm}
\section{Constrained Instance Reweighting}
\label{sec:formulation}
\vspace{-1mm}
We restrict ourselves to the classic supervised learning regime in this work (although it is possible to extend the method to other settings such as semi-supervised learning).
We use $x_i\in\reals^d$ to denote $i$th training example (iid sampled from distribution $P$) with its corresponding annotated label $y_i\in\{1,\ldots,K\}$, and use $\theta$ to denote model parameters. Let $L(x_i,y_i,\theta)>0$ be the loss for $i$th example, for  which we will use a shorthand of $L(x_i,\theta)$ for simplicity of notation. We assume that an unknown subset of the training examples has noisy labels (\ie, $y_i$ is not the true class). To address this label noise, we propose to reweight the training examples by assigning nonnegative weight $w_i$ to each example $x_i$. We propose the following population objective 
\vspace{-1mm}
\begin{align}
\inf_\theta \inf_Q \expect_{x\sim Q} L(x,\theta),\, \text{s.t. } D(Q,P)\leq\delta,
\label{eq:genopt_pop}
\vspace{-2mm}
\end{align}
\noindent where $D$ is a divergence of choice. This can be contrasted with the distributionally robust objective \cite{shapiro2017distributionally,levy2020large} which work with $\inf_\theta\sup_Q$ instead, with the goal of optimizing over worst case distributions. For practical purposes, we consider a finite-sample version of \eqref{eq:genopt_pop} as follows:
\vspace{-1mm}
\begin{equation}
\begin{split}
 \min_{\theta,w:w\geq 0,\sum_i w_i = 1} \sum_i w_i L(x_i,\theta),  \text{ s.t. } D(w,u)\leq \delta,
\end{split}
\label{eq:genopt}
\vspace{-2mm}
\end{equation}
\noindent where $D$ is now a divergence over discrete distributions and $u$ denotes the uniform distribution ($u_i=1/n$). To avoid maintaining weights over the full training set (which can be prohibitive at large scale), we propose to optimize \eqref{eq:genopt} separately for each minibatch. By optimizing  \eqref{eq:genopt}, we minimize an upper bound (in expectation) to \eqref{eq:genopt_pop} due to the following result.
\begin{theorem}
The finite-sample objective  $\min_{w\geq 0,\sum_i w_i = 1,D(w,u)\leq \delta} \sum_i w_i L(x_i,\theta)$ is an upper bound (in expectation) on the population objective $\inf_{Q:D(Q,P)\leq\delta} \expect_{x\sim Q} L(x,\theta)$.
\label{thm:upper_b}
\end{theorem}
We defer its proof to the Appendix. We leave establishing its sample convergence rate for future work and focus on the algorithmic aspects in the next sections. We refer to objective \eqref{eq:genopt} as Constrained Instance reWeighting or {\bf CIW}. 

\begin{table}[t]
\caption{\small Properties of some recent methods for addressing label noise.~{NEPM}: \emph{Does not} require extra persistent memory ($N$ is the number of training examples, $K$ is the number of classes), NEFP: \emph{No} extra forward pass needed (methods marked \xmark\, need an extra forward pass through the network), NSN: \emph{Does not} require training a separate network, STRM: Can be used in a streaming setting (often the case in large-scale production settings), NCL: \emph{Does not} require clean labels.}
\label{tab:compare}
\centering
\small
\begin{tabular}{c|ccccc}
\toprule
 \multirow{2}{*}{Methods} &  \multicolumn{5}{c}{Properties}\\[0.2ex]
&  \rot{40}{1em}{NEPM} & \rot{40}{1em}{NEFP} & \rot{40}{1em}{NSN} & \rot{40}{1em}{STRM} & \rot{40}{1em}{NCL} \\ \midrule
Mixup \citep{zhang2017mixup} & \cmark & \cmark & \cmark & \cmark & \cmark \\
APNL \citep{ma2020normalized} & \cmark & \cmark & \cmark & \cmark & \cmark \\
Bi-tempered \cite{bitempered} & \cmark & \cmark & \cmark & \cmark & \cmark \\
EG \cite{bar2021multiplicative,majidi2021exponentiated} & $\mathcal{O}(N)$ & \cmark & \cmark & \xmark & \cmark \\
Dynamic-Mixup \citep{arazo2019unsupervised} & $\mathcal{O}(N)$ & \xmark & \cmark & \xmark & \cmark \\
Divide-Mix \citep{Li2020DivideMix} &  $\mathcal{O}(N)$ & \xmark & \xmark & \xmark & \cmark \\ 
ELR \citep{liu2020early} & $\mathcal{O}(NK)$ & \cmark & \cmark & \xmark & \cmark \\ 
MentorNet \citep{jiang2018mentornet} & \cmark & \cmark & \xmark & \cmark & \xmark \\
LTRE \citep{ren2018learning} & \cmark & \xmark & \cmark & \cmark & \xmark \\
CICW (ours) & \cmark & \cmark & \cmark & \cmark & \cmark \\
\bottomrule
\end{tabular}
\vspace{-4mm}
\end{table}

\subsection{$f$-divergence}
\label{sub:fdiv}
\vspace{-1mm}
Let us take $D$ to be $f$-divergence \citep{sason2016f}. For simplicity, we work with the constraint $D(w,u)=\sum_i u_i f(w_i/u_i) \leq \delta$ where $f$ is a convex function with $f(1)=0$ (instead of a constraint on $D(u,w)$ which does not lead to simple closed form update rules). We can obtain the following update rule for the weights in this case. 
\begin{theorem}
For $f$-divergence constraint, \ie, $D(w,u)=\sum_i u_i f(w_i/u_i) \leq \delta$, the optimum weights for the problem \eqref{eq:genopt} for a fixed $\theta$ are given by
$w_i = u_i f'^{-1}\left(\frac{-L(x_i,\theta)-\mu+\nu_i}{\lambda}\right)$, where $u_i=1/n$ for uniform distribution, and $\lambda\geq 0$, $\mu$, and $\nu_i\geq 0$ are such that the constraints are satisfied. 
\label{thm:w_fdiv}
\end{theorem}
\remove{
Forming the Lagrangian for problem \eqref{eq:genopt}, we get
\begin{align}
\sum_i w_i L(x_i,\theta) + \lambda (D(w,u)-\delta) + \mu(\sum_i w_i - 1) -\sum \nu_i w_i,
\label{eq:gen_lag}
\end{align}
\noindent where $\lambda\geq 0, \nu_i\geq 0, \mu$ are the Lagrange multipliers. 
The dual function (for a fixed $\theta$) is given by
\begin{align}
h(\lambda,\mu,\nu) = \min_w \sum_i w_i L(x_i,\theta) + \lambda (D(w,u)-\delta)+ 
\mu(\sum_i w_i - 1) - \sum_i \nu_i w_i\, .
\label{eq:gen_dual}
\end{align}
Optimizing over $w$, the first order condition for optimality is
\begin{align}
\begin{split}
& L(x_i,\theta) + \lambda  f'\left(\frac{w_i}{u_i}\right) + \mu - \nu = 0\\
\Longrightarrow\, & w_i = u_i f'^{-1}\left(\frac{-L(x_i,\theta)-\mu+\nu_i}{\lambda}\right)= \frac{1}{n} f'^{-1}\left(\frac{-L(x_i,\theta)-\mu+\nu_i}{\lambda}\right)\, .
\end{split}
\label{eq:w_fdiv}
\end{align}}

We defer the proof to the Appendix. We adopt an alternating minimization approach for optimizing over $(w,\theta)$: fix $\theta$ and optimize for $w$ using the update of Theorem \ref{thm:w_fdiv}, then take a gradient step for model parameters $\theta$ while keeping the importance weights $w$ fixed. As we define the problem \eqref{eq:genopt} over a single minibatch, there is no extra overhead of maintaining the importance weights over  the entire training data or across the training iterations. 
We can also obtain closed form solutions for the importance weights when $D$ is taken to be in the family of Bregman divergence~\citep{bregman}. We provide more details on this in the Appendix. 
\remove{
\subsection{Bregman divergence}
\label{sub:bregman}
The Bregman divergence using the convex function $F$ is defined as
\[
D(u, w) = F(u) - F(w) - \nabla F(w)\cdot (u - w)\, .
\]
For convenience, we use $f \coloneqq \nabla F$. We again work with inverse Bregman divergence $D(w,u)$ to obtain  closed form updates. We have same dual function as in \eqref{eq:gen_dual}. 
First order optimality condition for $w$ (for a fixed $\theta$) is given by 
\begin{align}
L(\theta) + \lambda (f(w) - f(u)) + \mu\,1 - \nu = 0 \Longrightarrow w = f^{-1}\left(f(u) - \frac{L(\theta) + \mu1 - \nu}{\lambda}\right)\,,
\end{align}
where $L(\theta)$, and $\nu$ are the vectors of losses and Lagrange multipliers, respectively, and $\mu1$ is a vector with each entry equal to $\mu$. The Lagrange multipliers $\lambda$, $\mu$ and $\nu_i$ are such that the constraints are satisfied. 
}

\vspace{-1mm}
\subsection{Some Special Cases of Divergences}
\vspace{-1mm}
We now consider some special cases of commonly used divergences. 

\subsubsection{KL Divergence}
\label{sub:kl}
\kl-divergence belongs to both $f$-divergence and Bregman divergence family, and is given by $D(w,u) = \kl(w,u)=\sum_i w_i \log \frac{w_i}{u_i}$. It can be obtained by taking the generating function $f(t)=t\log t$ in $f$-divergence, which in turn implies $f'^{-1}(t)=e^{t-1}$.
Since $f'^{-1}(t)=e^{t-1}>0$ for all finite $t$, all weights are non-zero and we will have $\nu_i=0$. 
Hence, using Theorem \ref{thm:w_fdiv}, the weights are given by
\vspace{-1mm}
\begin{align}
w_i = \frac{1}{n} \exp\left(-\frac{L(x_i,\theta)+\mu}{\lambda}-1\right)  = \frac{\exp\left(-\frac{L(x_i,\theta)}{\lambda}\right)}{\sum_j \exp\left(-\frac{L(x_j,\theta)}{\lambda}\right)}\, .
\label{eq:w_kldiv}
\end{align}
The second equality above is obtained by using the fact that $\sum_i w_i = 1$.
The Lagrange multiplier $\lambda$ is such that the constraint $D(w,u)\leq\delta$ is active. In our experiments, we use $\lambda$ as the tunable hyperparameter instead of $\delta$. 
\vspace{-1mm}
\subsubsection{Reverse-KL Divergence}
\label{sub:revkl}
\vspace{-1mm}
Reverse-\kl~divergence also belongs to both $f$-divergence and Bregman divergence family, and is given by $D(w,u) = \kl(u,w)=\sum_i u_i \log \frac{u_i}{w_i}$. It can be obtained by taking the generating function $f(t)=-\log t$ in $f$-divergence, which in turn implies $f'^{-1}(t)=-\frac{1}{t}$. 
Since a zero weight will result in unbounded reverse-\kl~divergence and violate the constraint, all weights have to be positive and we will have $\nu_i=0$. 
Hence, using Theorem \ref{thm:w_fdiv}, the weights are given by
\vspace{-1mm}
\begin{align}
w_i = \frac{1}{n} \frac{\lambda}{L(x_i,\theta)+\mu} = \frac{1/(L(x_i,\theta)+\mu)}{\sum_j 1/(L(x_j,\theta)+\mu)}\, .
\label{eq:w_revkl}
\end{align}
Again, the second equality above is obtained by using the fact that $\sum_i w_i = 1$.
The Lagrange multiplier $\mu$ is such that the constraint $D(w,u)\leq\delta$ is active. In our experiments, we use $\mu$ as the tunable hyperparameter instead of $\delta$.

\begin{figure*}[t!]
\vspace{-3mm}
\begin{center}
    \subfigure{\includegraphics[width=0.27\linewidth]{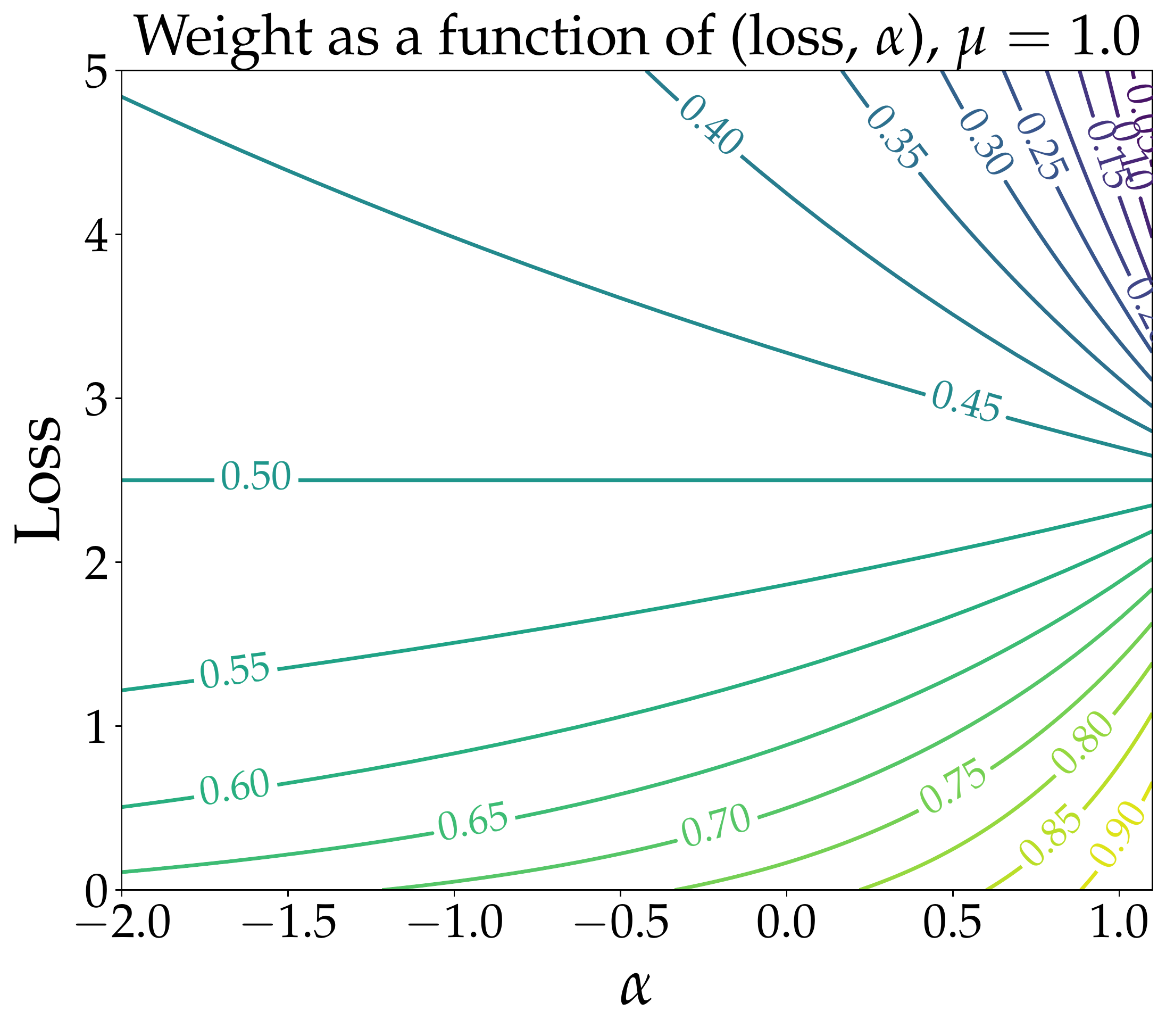}}\,\,  
    \subfigure{\includegraphics[width=0.27\linewidth]{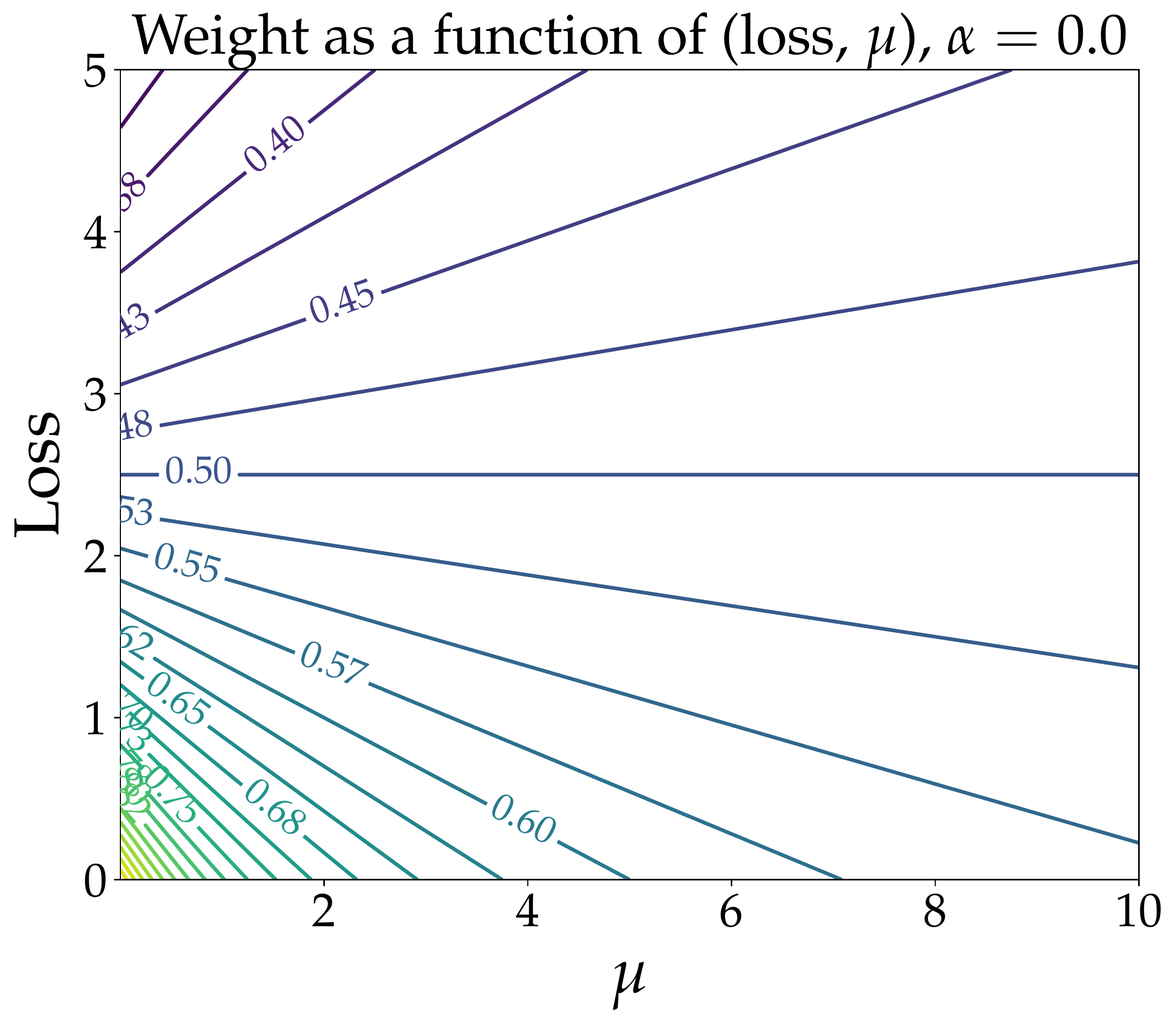}}
    \subfigure{\includegraphics[width=0.27\linewidth]{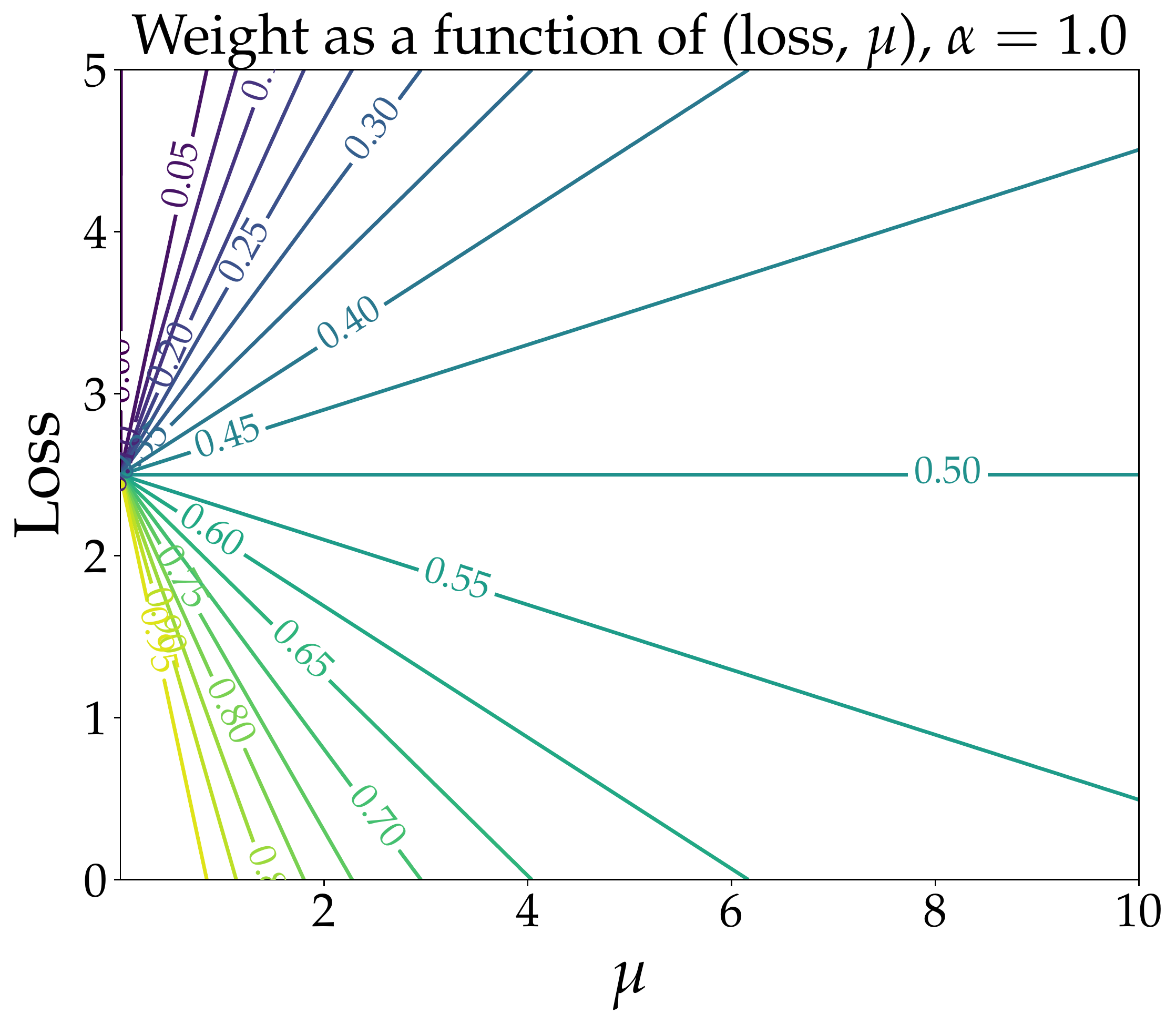}}
    \vspace{-5mm}
    \caption{\small Level sets of the weight of an example as a function of its loss in a batch size of $2$, where the loss of the other example is fixed to $2.5$. The horizontal line passing through the middle corresponds to a value of loss equal to $2.5$, thus having instance weight $=0.5$. Points above this line (with loss $> 2.5$) induce weights $< 0.5$ and vice versa. (a) Effect of different $\alpha$ for a fixed $\mu=1$. The level sets are \emph{asymmetric} across the center line. Also, the level sets become denser along a vertical slice for larger $\alpha$, as the distribution becomes less smooth. (b) Effect of $\mu$ for a fixed $\alpha=1$. The level sets are again asymmetric across the center line and become denser for smaller $\mu$. (c) Similar plot for $\alpha = 1$. The change in level sets is more rapid as the distribution has shorter tail.}
    \label{fig:w_contours}
    \end{center}
    \vspace{-3mm}
\end{figure*}

\vspace{-1mm}
\subsubsection{Other $\alpha$-divergences and Generalized Softmax}
\label{sub:alpha}
\vspace{-1mm}
$\alpha$-divergence parameterized by $\alpha \in \R$ is a class of $f$-divergence which is commonly used in machine learning \citep{amari2009alpha}. The $\alpha$-divergence is induced by the generating convex function $f_{\alpha}(t) = \frac{1}{\alpha(1-\alpha)} (t - t^\alpha)$ for $\alpha \in \R \setminus \{0, 1\}$, $f_{0}(t)=-\log t$ and $f_{1}(t)=t\log t$. 
\remove{
\begin{equation}
    f_{\alpha}(t) = \begin{cases}
    \frac{1}{\alpha(1-\alpha)} (t - t^\alpha) & \alpha \in \R \setminus \{0, 1\}\\
    -\log t & \alpha = 0\\
    t\log t & \alpha = 1
    \end{cases}\, .
\end{equation}}

$\alpha$-divergence recovers many well-know divergences for different values of $\alpha$, including Neyman-$\chi^2$ ($\alpha = -1$), Reverse-KL ($\alpha=0$), Hellinger ($\alpha=0.5$), KL ($\alpha=1$, and Pearson-$\chi^2$ ($\alpha = 2$). As we already handle the case of $\alpha = 0$ and $\alpha=1$ in the previous sections, we focus on $\alpha \in \R \setminus \{0, 1\}$. We have
$f'(t) = -\frac{1}{1-\alpha}\, t^{\alpha-1}$ which yields the inverse function $f'^{-1}(t) = \big((\alpha - 1)\,t \big)^{\frac{1}{\alpha - 1}}$. %
The constraint $w_i \geq 0$ may become active for certain values of $\alpha > 1$, thus the Lagrange multiplier $\nu_i$ is positive in Theorem \ref{thm:w_fdiv} causing the weight to be zero (\ie, $\nu_i=L(x_i,\theta)+\mu$). As a result, the weights are given by 
\vspace{-2mm}
\begin{equation}
\label{eq:w_alphadiv_raw}
w_i = \frac{[(1 - \alpha)L(x_i,\theta)+\mu]_+^{1/(\alpha - 1)}}{\sum_j [(1 - \alpha)L(x_j,\theta)+\mu]_+^{1/(\alpha - 1)}}\,,\quad \alpha \neq 1\,, 
\end{equation}
where $[\,\cdot\,]_+ = \max(\,\cdot\,, 0)$. The limiting case of $\alpha=1$ which corresponds to KL divergence is given in Eq.~\eqref{eq:w_kldiv}. Eq.~\eqref{eq:w_alphadiv_raw} can alternatively be viewed in the form of a generalized softmax function. Using the definition of the generalized exponential function $\exp_s(t) \coloneqq [1 + (1 - s) t]_+^{\frac{1}{1-s}},\, s\in \R \setminus \{1\}$ and $\exp_1(t) = \exp(t)$ as defined in~\citep{texp1}, we can also write the weights as
$w_i = \frac{\exp_{(2-\alpha)}\!\left(-\frac{L(x_i,\theta)}{\mu}\right)}{\sum_j \exp_{(2-\alpha)}\!\left(-\frac{L(x_j,\theta)}{\mu}\right)}$. 
\remove{
\begin{equation}
    w_i = \frac{\exp_{(2-\alpha)}\!\left(-\frac{L(x_i,\theta)}{\mu}\right)}{\sum_j \exp_{(2-\alpha)}\!\left(-\frac{L(x_j,\theta)}{\mu}\right)}\, .
\end{equation}
}
As we decrease $\alpha$, the distribution of weights will have heavier tails (\ie, the difference between the weights for large and small losses will be less, resulting in a flatter distribution). Similarly for a fixed $\alpha$, we will see heavier tails with increasing $\mu$. We illustrate this behavior in Figure
\ref{fig:w_contours}. 

\begin{figure*}[t!]
\vspace{-0.1cm}
\begin{center}
 \subfigure[Baseline boundary at epoch 6]{\includegraphics[width=0.19\textwidth]{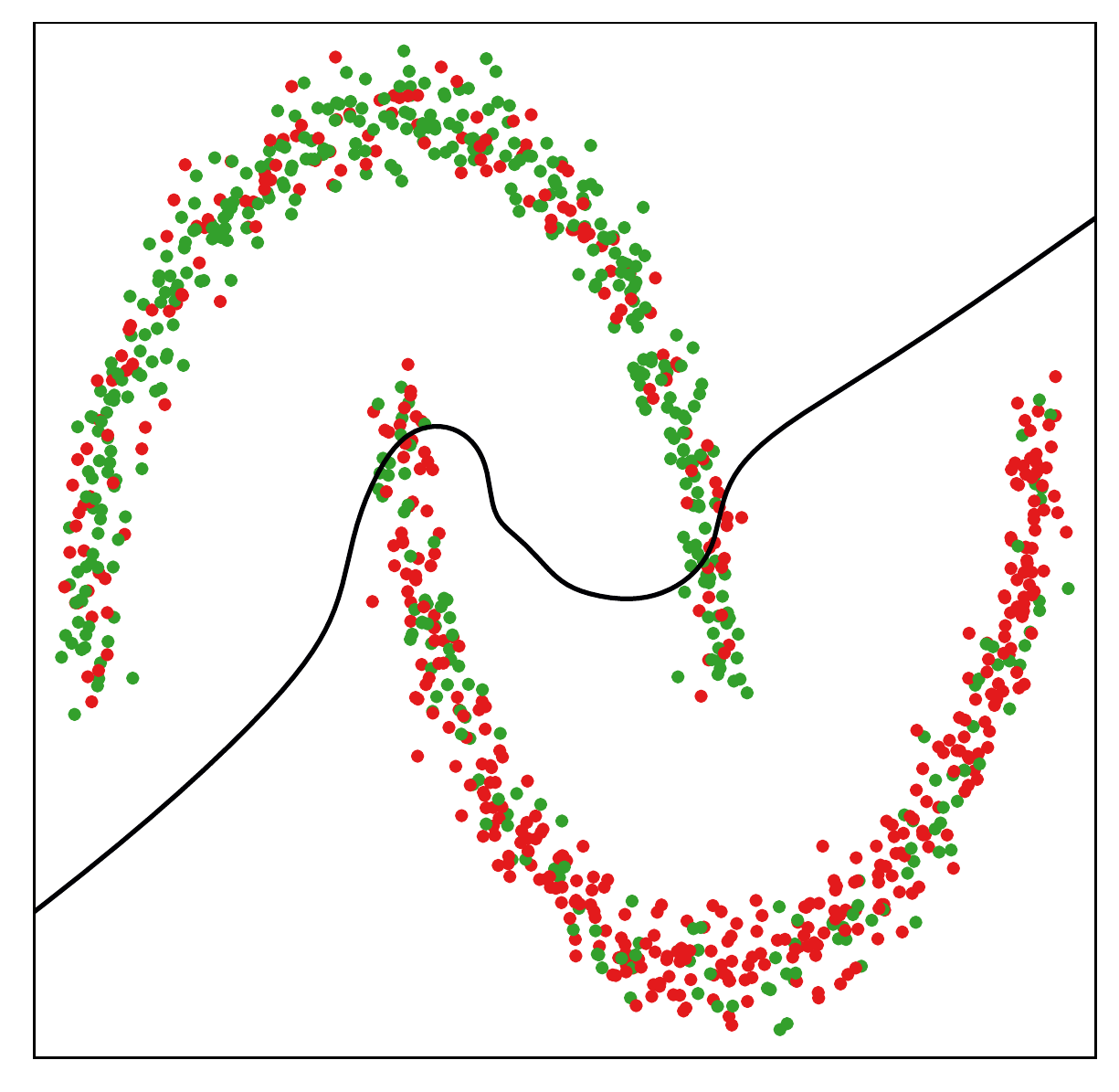}}%
 \subfigure[Baseline boundary at epoch 20]{\includegraphics[width=0.19\textwidth]{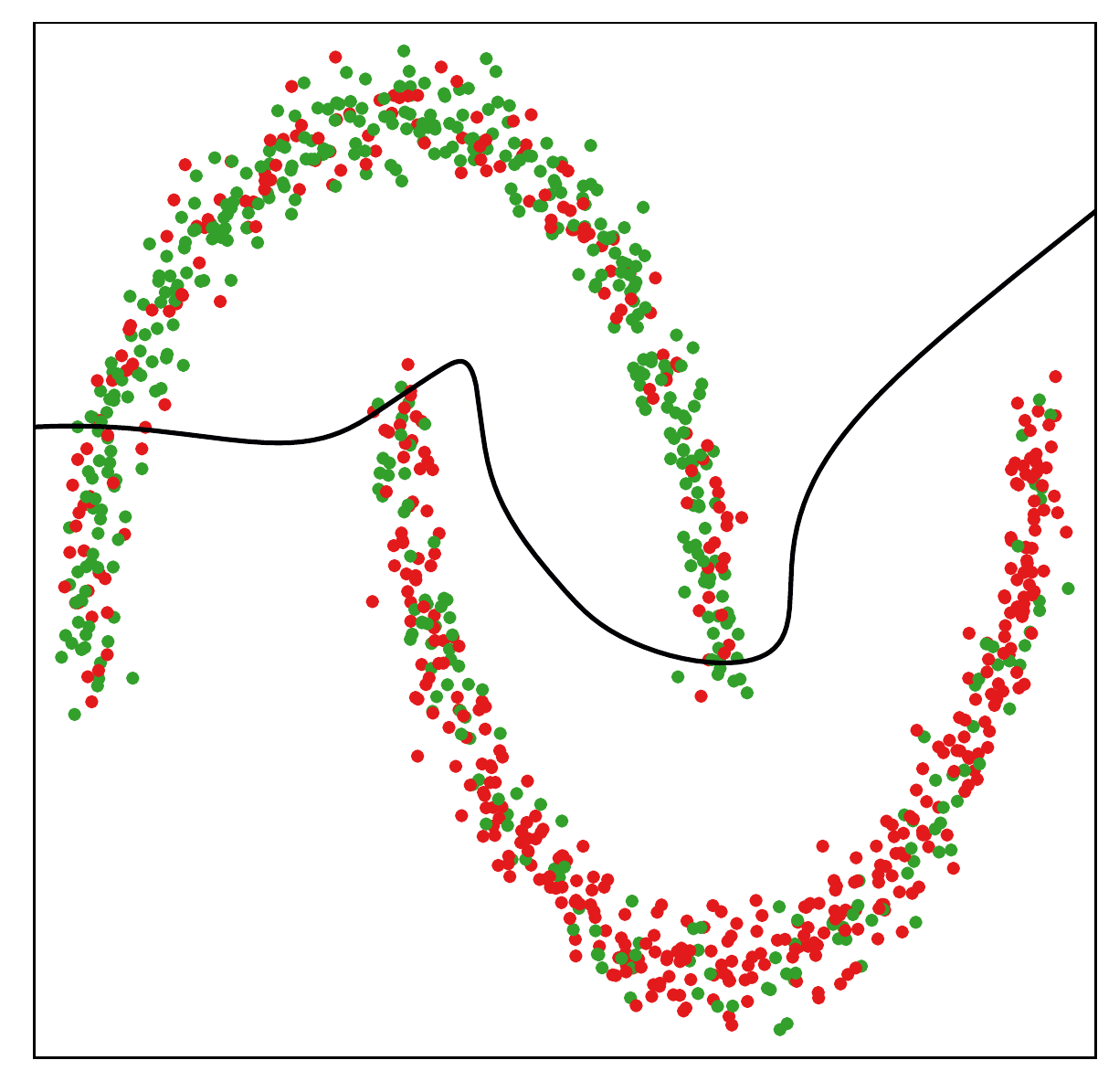}}%
 \subfigure[A mini-batch of examples at epoch 6]{\includegraphics[width=0.19\textwidth]{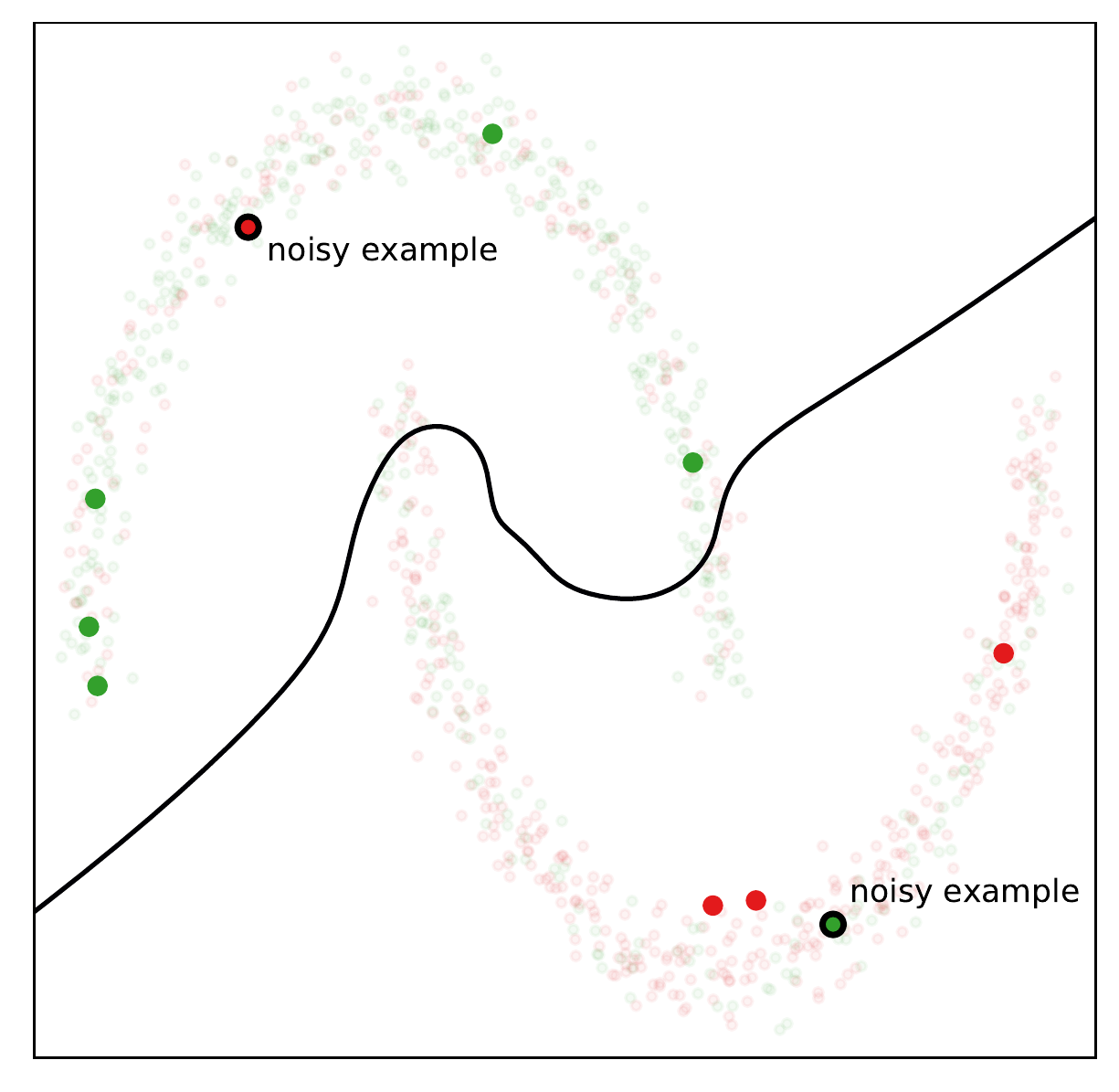}}%
 \subfigure[The same mini-batch reweighted by Eq.~\eqref{eq:w_alphadiv_raw}]{\includegraphics[width=0.19\textwidth]{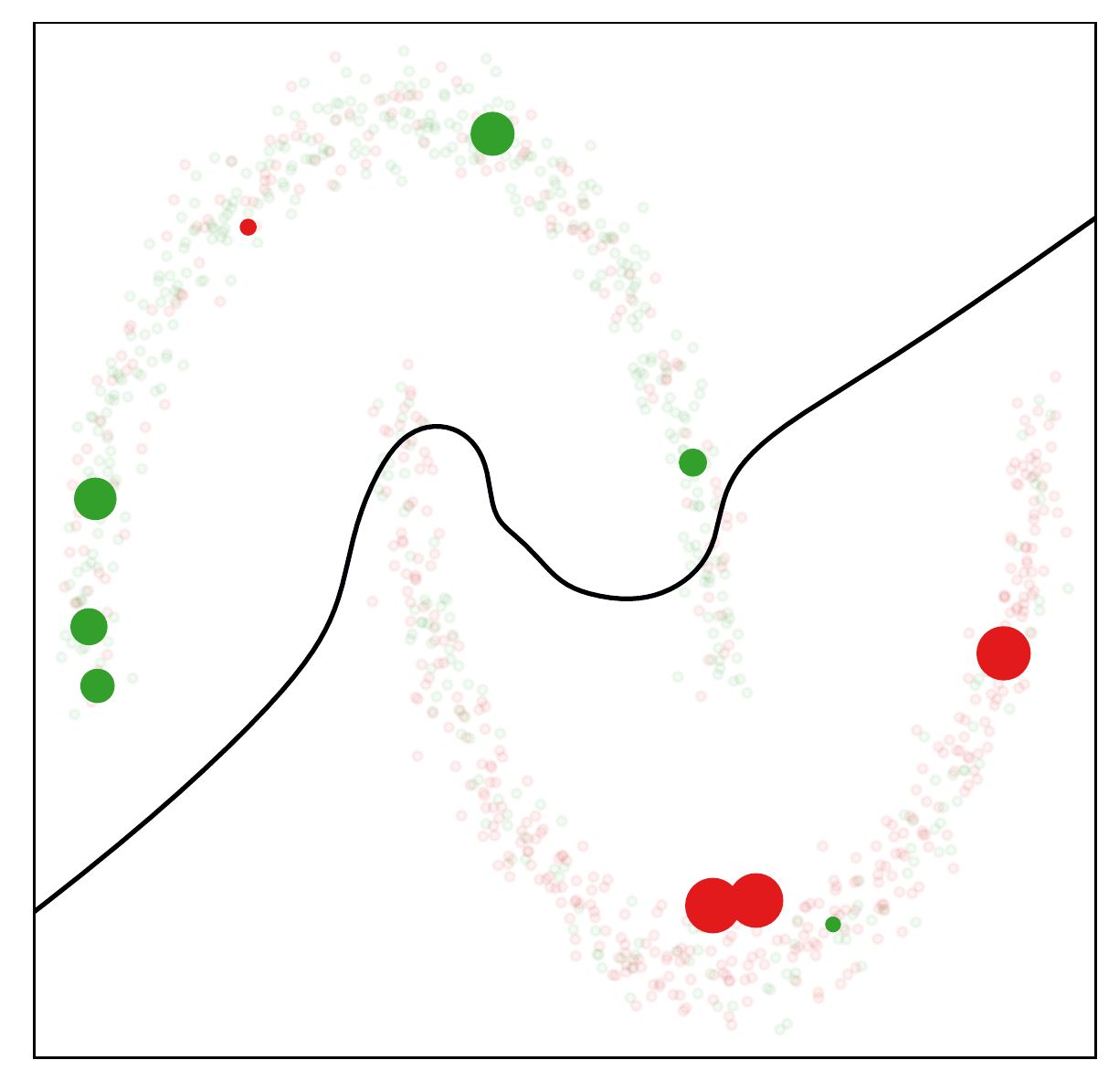}}%
 \subfigure[CIW boundary at epoch 20]{\includegraphics[width=0.19\textwidth]{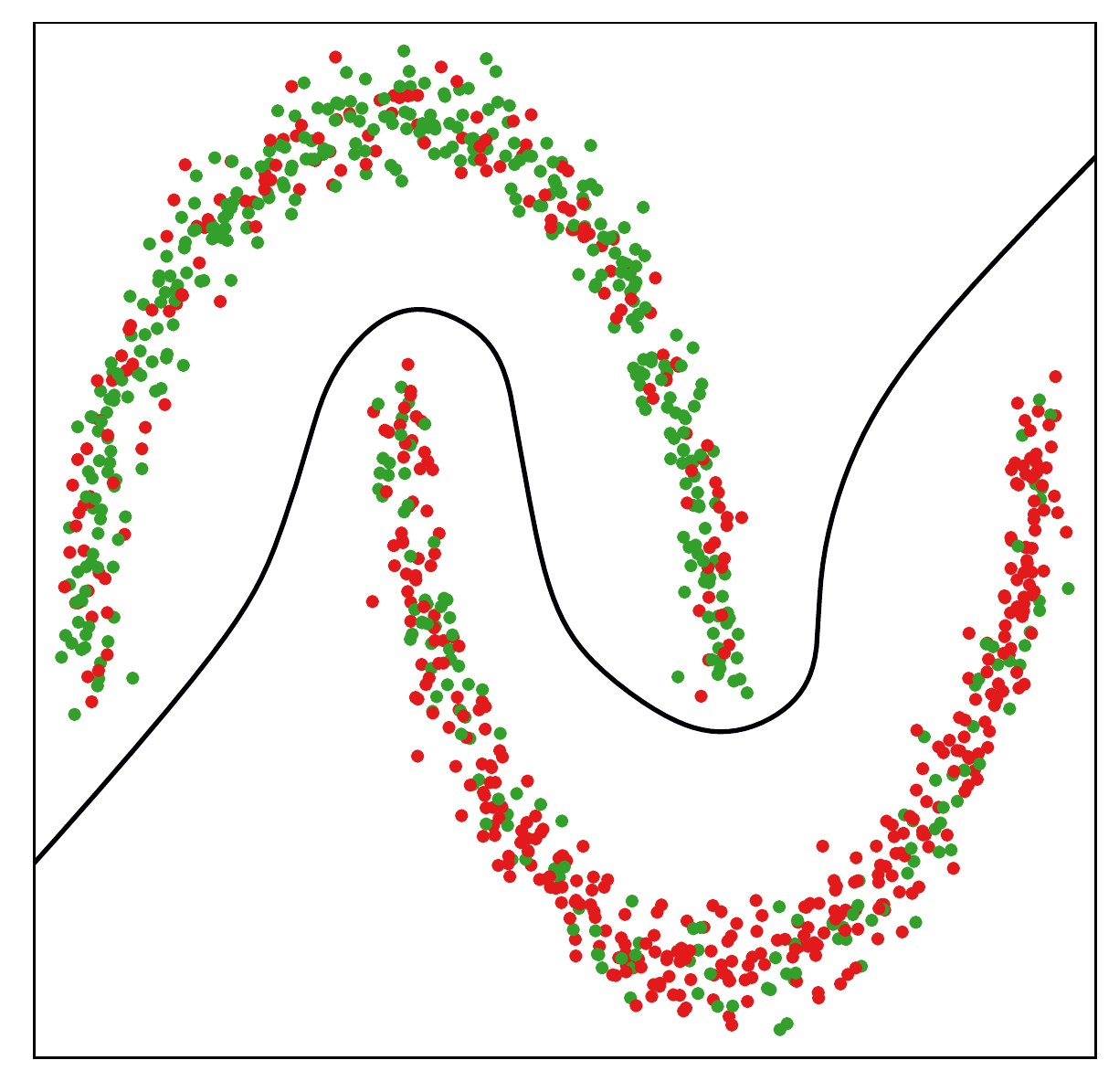}}
\vspace{-5mm}
\caption{\small Illustration of our proposed Constrained Importance Reweighting (CIW) method on a two-layer neural network trained on the noisy Two Moons dataset: %
(a) The decision boundary of the baseline model after 6 epochs. %
(b) The decision boundary of the baseline model after 20 epochs. The large loss of the misclassified noisy examples causes the model to eventually overfit to noise. c) A random mini-batch of examples at epoch 6. The baseline model treats these examples as equally important. (d) The same mini-batch of examples reweighted by our proposed approach (with size of each example indicating its importance). (e) By activating the proposed instance reweighting at epoch 6, the model is able to %
fit well to the geometry of the data at epoch 20.}
\label{fig:moons}
\end{center}
\vspace{-0.5cm}
\end{figure*}

We also show the effect of our instance reweighting approach in a toy noisy binary classification setting in two dimensions. We use a two-layer fully-connected neural network with $\tanh$ activations and 10 and 20 hidden layers, respectively. The model is trained on 1000 samples from the Two Moons dataset\footnote{\url{https://scikit-learn.org/stable/modules/generated/sklearn.datasets.make_moons.html}} with 30\% random flip label noise. Figure~\ref{fig:moons} visualizes the decision boundary of the baseline model, trained with the CE loss, as well as with the reweighted loss via Eq.~\eqref{eq:w_alphadiv_raw} (using $\alpha=0.5$ and $\mu=0.5$). More details are deferred to the Appendix. Our importance reweighting approach is able to successfully rectify the decision boundary by emphasizing on the clean examples in each batch while downweighting the \mbox{noisy ones.}
\vspace{-1mm}
\section{Constrained Class Reweighting}
\vspace{-1mm}
\label{sub:cicw}
\vspace{-1mm}
Instance reweighting presented in the earlier sections assigns high weights to instances with lower losses while not deviating far from a uniform distribution over instances.
In this section, we extend this intuition to assign importance weights over all possible class labels. For the mislabeled examples, it is reasonable to assign non-zero weights to classes that could potentially be the true label. Let us denote by $L_j(x_i,\theta)$ the loss for example $x_i$ with the assumption that the true label is class $j$, \ie, $L_j(x_i,\theta)\coloneqq L(x_i,j,\theta)$ (note that we used $L(x_i,\theta)$ to denote $L(x_i,y_i,\theta)$ in the earlier sections, where $y_i$ was the annotated label). We now consider the  optimization problem:
\vspace{-2mm}
\begin{equation}
\setlength\belowdisplayskip{1pt}
\begin{split}
\min_{w,v,\theta}  \sum_i w_i \left[\sum_j v_{ij} L_j(x_i,\theta)\right],
\text{s.t. } & D_1(u,w)\leq \delta, \sum_i w_i = 1, w_i \geq 0\,, \\
& D_2(e_{i},v_i)\leq \gamma\, \forall i, \sum_j v_{ij}=1, v_{ij} \geq 0\, \forall i\,,
\end{split}
\label{prob:class_w}
\end{equation}
\noindent where index $i$ runs over the examples in the minibatch, index $j$ runs over all the classes, $u$ is the uniform distribution over examples, $v_{ij}$ is the weight for class $j$ for $i$th example, $e_i$ is the one-hot vector with $1$ in the position of annotated class of $i$th example (\ie, $y_i$). 
Since the inner problem for every example is independent of others, we first solve each  inner problem independently to get class weights $v$ and fix them before computing instance weights $w$. %
We refer to objective \eqref{prob:class_w} as Constrained Instance and Class reWeighting or {\bf CICW}. 
Next we consider some special cases for divergence $D_2$ and derive updates for class weights.
\vspace{-1mm}
\subsection{Total Variation}
\label{sub:cls_tv}
\vspace{-2mm}
Taking $D_2$ to be the total variation distance will result in a linear program in $v$ with solution lying on a vertex. We show the following result in this case. 
\remove{
We can rewrite the optimization problem \eqref{prob:class_w} in $v$ (omitting the example index $i$) as 
\begin{equation}
\begin{split}
&\min_v \sum_j v_{j} L_j(x,\theta),\\& \text{ s.t. }  \lVert e - v\rVert_1 \leq \gamma, \sum_j v_{j}=1, v_{j} \geq 0\,,
\end{split}
\label{prob:class_w_tv}
\end{equation}
where $e$ is a one-hot vector with $e_y=1$ and $e_j=0\, \forall j\neq y$.
}
\remove{
We can write $\lVert e - v\rVert_1=(1-v_{y}) + \sum_{j\neq y} v_{j}$, where $y$ is the label of example $x$. Since $\sum_j v_{j}=1$, the constraint $\lVert e - v\rVert_1\leq \gamma$ transforms into $v_y\geq 1-\gamma/2$ and $\sum_{j\neq y} v_j \leq \gamma/2$. Substituting it in \eqref{prob:class_w_tv} yields
\begin{align}
\min_v \sum_{j} v_{j} L_j(x,\theta),\quad \text{s.t. }  v_y\geq 1-\gamma/2, \sum_{j\neq y}v_j \leq \gamma/2, \sum_j v_j = 1, v_{j} \geq 0
\label{prob:class_w_tv2}
\end{align}
If $y=\argmin_j L_j(x,\theta)$, the solution of the linear program in \eqref{prob:class_w_tv2} will be the one-hot vector with $1$ at the position $y$. For the case $\hat{y}=\argmin_{j} L_j(x,\theta)\neq y$, the solution will be a two-hot vector with $1-\gamma/2$ at the position $y$ and $\gamma/2$ at the position $\hat{y}$. 
}
\begin{theorem}
\setlength\belowdisplayskip{0pt}
For $D_2(e,v)=\lVert e-v\rVert_1$ (total variation distance), the objective \eqref{prob:class_w} reduces to
\begin{equation*}
\setlength\belowdisplayskip{0pt}
\begin{split}
&\min_{w,\theta} \sum_i w_i \left[(1-\gamma/2) L_{y_i}(x_i,\theta) + \gamma/2 L_{\hat{y_i}}(x_i,\theta)\right], \\ &\text{ s.t. } D_1(u,w)\leq \delta, \sum_i w_i = 1, w_i \geq 0\,,
\end{split}
\label{prob:class_w_tv3}
\end{equation*}
where $\hat{y_i}$ denotes the class with lowest loss.
\label{thm:v_totalvar}
\end{theorem}
We defer the proof of this fact to the Appendix. 
We note that this is same as static bootstrapping  \citep{reed2014training,arazo2019unsupervised} which was earlier proposed for label noise in a rather heuristic manner. Theorem \ref{thm:v_totalvar} shows that it can justified in a principled manner from the point of view of constrained optimization over class weights. 
It is possible to use a per-instance $\gamma_i$, perhaps making it a function of the class losses (\ie, $\gamma_i=g(L_1(x_i,\theta),\ldots,L_K(x_i,\theta))$ for some function $g$), but we work with a $\gamma$ that is globally fixed for all instances in our experiments for the sake of simplicity. Later, we introduce a Mixup variant that utilizes CIW for dynamically adjusting $\gamma_i$ per example.

{\bf Other divergences that result in similar solution for class weights.~} The effective \emph{inner loss} in \eqref{prob:class_w_tv3} is a convex combination of the two losses: loss of the annotated class $y_i$ and loss of the predicted class $\hat{y}_i$. It can be shown that similar weighting of the two losses, with weights given by $(1-g(\gamma))$ and $g(\gamma)$ for some nonnegative function $g$, are obtained if we take $D_2(\cdot,\cdot)$ to be $\ell_\infty$ distance, reverse-\kl~divergence, or reverse $f$-divergence. We provide the proof of this fact in the Appendix. 

\vspace{-2mm}
\subsection{$\ell_2$-distance}
\vspace{-1mm}
\label{sub:cls_l2}
We now take $D_2$ to be the squared $\ell_2$-distance  and consider the following problem:
\begin{equation}
\setlength\abovedisplayskip{2pt}
\setlength\belowdisplayskip{0pt}
\label{prob:class_w_l2}
\begin{split}
 \min_v \sum_j v_{j} L_j(x,\theta), & \text{ s.t. } \lVert e - v\rVert_2^2 \leq \gamma, \sum_j v_{j}=1, v_{j} \geq 0
\end{split}
\vspace{-6mm}
\end{equation}
\remove{
Lagrangian of the above problem is given by $\mcal{L}(\theta,\lambda,\mu,\nu)=\sum_j v_{j} L_j(x,\theta) + \lambda/2 (\lVert v - e\rVert^2 - \gamma) + \mu(\sum_j v_j -1) - \sum_j \nu_j v_j$, 
where $\lambda\geq 0$ and $\nu_j\geq 0$. Minimizing it w.r.t. $\theta$ gives
\begin{align}
v_j = \frac{-1}{\lambda}(L_j(x,\theta)-\nu_j+\mu) + e_j
\label{eq:v_l2_cls}
\end{align}
Since $\sum_j v_j=1$, we get $\mu=-\frac{1}{k}\sum_j(L_j(x,\theta)-\nu_j)$, where $k$ is the number of classes, $e_y=1$ and $e_j=0$ for $j\neq y$. 
When $y=\argmin_j L_j(x,\theta)$, the solution of \eqref{prob:class_w_l2} will be the one-hot vector with $1$ at the position $y$ and the constraint $\lVert e-v\rVert_2^2\leq\gamma$ will be inactive. For the case $\hat{y}=\argmin_{j} L_j(x,\theta)\neq y$, this constraint will be active (\ie, $\lVert e-v\rVert_2^2=\gamma$) and we will have
\begin{align}
\lambda^2 = \frac{1}{\gamma}\sum_j (L_j(x,\theta)-\nu_j+\mu)^2
\label{eq:lambda_l2_cls}
\end{align}
For indices $j$ s.t. $v_j>0$, we will have the dual variables $\nu_j=0$. In practice, we would like to avoid setting a $\gamma$ so large that it assigns a zero weight to the annotated class (\ie, $v_y$ should be positive). In this case,
for indices $j\neq y$ s.t. $v_j=0$, we will have $L_j(x,\theta)-\nu_j=-\mu$. Since $\mu$ is the mean of all $\{\nu_j-L_j(x,\theta)\}_{j=1}^k$ and a subset of these are equal to $\mu$ (\ie, indices $Z=\{j: v_j=0\}$), $\mu$ should also be the mean of the complementary subset $\bar{Z}$, \ie,
\begin{align}
\mu=\frac{1}{|\bar{Z}|}\sum_{j\in \bar{Z}} (\nu_j-L_j(x,\theta))= -\frac{1}{|\bar{Z}|}\sum_{j\in \bar{Z}} L_j(x,\theta).
\label{eq:mu_l2_cls}
\end{align}
For the non-zero indices in $\bar{Z}\setminus y$, we will have $L_j(x,\theta)<-\mu$ (as $v_j>0$). For $v_y>0$, we have $L_y(x,\theta)<\lambda-\mu$, 
and for $v_y=0$ we will have $L_y=\lambda+\nu_y-\mu\geq\lambda-\mu$. }
The solution to this can be obtained by sorting the losses $\{L_j(x,\theta)\}_{j=1}^k$ in ascending order and doing a search for the nonzero indices of $v$ over the possible solution set $S=\{[m]\cup\tilde{y}\}_{1\leq m<\tilde{y}}$, where $[m]=\{1,\ldots,m\}$, and $\tilde{y}$ denotes the index where the loss $L_y(x,\theta)$ falls in this ranking of losses. The number of candidate solutions are $\tilde{y}$ and the correct solution can be identified by checking certain conditions. We provide the complete proof for this in the Appendix. 
A first heuristic that often provided correct solutions in our experiments was to compute the mean $\tilde{\mu}$ of the losses $\{L_j(x,\theta):L_j(x,\theta)\leq L_y(x,\theta)\}$ and set the nonzero indices of $v$ to be $\{j:L_j(x,\theta)<\tilde{\mu}\}\cup \{y\}$. 
Since the problem \eqref{prob:class_w_l2} is convex, we can also use a convex solver to solve for  $v$. 
However, our analysis gives  an interesting insight behind the working of the $\ell_2$-distance constraint, \ie, it spreads the mass of the weight vector $v$ more broadly to classes with low losses than the total variation distance which only allots the mass to the class with least loss.

\vspace{-3mm}
\section{Mixup With Importance Weights}
\label{subsec:mixup}
\vspace{-1mm}
Mixup \citep{zhang2017mixup} has been shown to work well in the presence of label noise \citep{arazo2019unsupervised}. However, vanilla Mixup uses randomly sampled weights to combine the two images and their labels which can be suboptimal in the presence label noise. 
We propose two ways to combine Mixup with the instance weights obtained using our method that further provide significant empirical gains against label noise:

{\bf (i) Using importance weights for mixing (IW-Mix).~} Let $X\in\reals^{n\times d}$ denote a training minibatch of $n$ examples. The $i$th example of the mixed up minibatch $X^{(m)}\in\reals^{n\times d}$ is given by $X^{(m)}_{i:}=(w_i X_{i:} + \tilde{w}_i \tilde{X}_{i:})/(w_i+\tilde{w}_i)$ where $\tilde{w}$ and $\tilde{X}$ are obtained by applying same random permutation $\mcal{P}$ to both $w$ and the rows of $X$. The labels are also obtained by same mixing proportions. If $Y\in\reals^{n\times K}$ is the one-hot label matrix, then the $i$th mixed up label is given by $Y^{(m)}_{i:}=(w_i Y_{i:} + \tilde{w}_i \tilde{Y}_{i:})/(w_i+\tilde{w}_i)$, where $\tilde{Y}$ is the obtained by applying the same permutation $\mcal{P}$ to $Y$. 

{\bf (ii) Using importance weights for both sampling and mixing (SIW-Mix).~} In this variant, we importance sample the example indices (with replacement) in the minibatch using instance weights $w$ as probabilities. For the importance sampled indices $I$, let $\tilde{w}=w[I]$, $\tilde{X}=X[I,:]$ and $\tilde{Y}=Y[I,:]$. 
We then construct the mixed up minibatch as earlier, \ie, $X^{(m)}_{i:}=(w_i X_{i:} + \tilde{w}_i \tilde{X}_{i:})/(w_i+\tilde{w}_i)$ and $Y^{(m)}_{i:}=(w_i Y_{i:} + \tilde{w}_i \tilde{Y}_{i:})/(w_i+\tilde{w}_i)$.  

{\bf Using the mixed up batch.~} There are two ways we can use the mixed up batch during training: (i) {\bf Mixup-base:} Simply compute the base loss $L(x^{(m)}_i,y^{(m)}_i,\theta)$ for each mixed up example $(x^{(m)}_i,y^{(m)}_i)$ and use the average loss $\frac{1}{n}\sum_i L(x^{(m)}_i,y^{(m)}_i,\theta)$ for backpropagation, (ii) {\bf Mixup-reweight:} Use the losses for mixed up examples $\{L(x^{(m)}_i,y^{(m)}_i,\theta)\}_{i=1}^n$ as our base losses and plug them into the problem \eqref{prob:class_w},  recompute the instance and class weights for the mixed up examples, and use the final reweighted loss for backprogation. We experiment with both these strategies. 
Since earlier derived solutions %
to the problem \eqref{prob:class_w},
when $D_2$ is total-variation or squared-$\ell_2$ distance, 
hold only for one-hot vectors $e$ and mixed up labels are two-hot, we use a Python based convex solver, CVXPY \citep{diamond2016cvxpy,agrawal2018rewriting}, %
for solving for the class reweighting $v$ in the case of Mixup-reweight strategy. %
We collectively refer to these combinations of CICW and Mixup as {\bf CICW-M}.

{\bf Dynamic label-smoothed Mixup with our importance weights.}
Inspired by the dynamic label smoothing approach in~\citep{arazo2019unsupervised}, we also consider using our CIW instance weights to smoothen the labels of the batch of examples before applying vanilla Mixup. This is equivalent to setting $\gamma$ in the CICW objective \eqref{prob:class_w} as a function of  instance weights $w$ and use $D_2$ to be total variation distance. In this approach, after calculating $w$ for a batch of examples, we calculate the normalized weight $\hat{w}_i = \frac{w_i - \min_j w_j}{\max_j w_j - \min_j w_j}$ for the $i$th example. This weight is then used for smoothing the example label as $\bar{y}_i = \hat{w}_i y_i + (1 - \hat{w}_i) z_i$, where $z_i$ is the one-hot label corresponding to the predicted class for the $i$th example. Finally, we apply vanilla Mixup on the set of examples $\{(x_i, \bar{y}_i)\}$. We refer to this as {\bf Dyn-CICW-M}.

\vspace{-3mm}   
\section{Related Work}
\label{sec:related}
\vspace{-3mm}
There has been a plethora of research on handling noisy labels in supervised learning. These methods span from improving the data representation to modifying the network architecture as well as the optimization techniques. Some  approaches aim to either directly estimate the clean labels from the corrupted labels~\citep{hendrycks2018using,extendedT,collier2021correlated}, or estimate the noise rates to get the density ratio \citep{liu2015classification}.
Other approaches consider adding regularizer terms to avoid overfitting to noisy examples~\citep{azadi2015auxiliary,miyato2018virtual,foret2021sharpnessaware,liu2020early}. These methods also include label smoothing~\citep{szegedy2016rethinking,lukasik2020does} and data augmentation techniques such as mixup~\citep{zhang2017mixup} and dynamic mixup~\citep{arazo2019unsupervised} to control the network's confidence on mixed corrupted examples. Example reweighting has been recently considered to combat the noisy examples that incur comparatively larger loss values, thus causing the network to overfit to these examples~\citep{jiang2018mentornet,ren2018learning,bar2021multiplicative,majidi2021exponentiated}. However these methods need to either maintain weights over full train set \citep{bar2021multiplicative,majidi2021exponentiated}, or need a clean subset of data to learn the rweighting \citep{ren2018learning,jiang2018mentornet} and train a separate network \citep{jiang2018mentornet}.  Generalized loss functions such as generalized bi-tempered cross entropy loss~\citep{bitempered}, normalized loss functions~\citep{ma2020normalized}, and peer loss~\citep{peerloss} achieve this by bounding the contribution of each example or punishing over-confidence. Semi-supervised techniques have also been employed to improve generalization~\citep{nguyen2019self,Li2020DivideMix}.
Several approaches have been proposed to adjust the network architecture~\citep{xiao2015learning,Goldberger17Training,han2018co,wei2020combating} as well. Methods such as~\citep{li2020gradient,drory2020resistance} analyze the effect of different factors for the robustness of the network. Recent approaches such as~\citep{liu2020early,han20c,xia2021robust} analyze the memorization effect of the network. Curriculum learning is another approach which has shown promise to reduce the effect of label noise~\citep{jiang2018mentornet,saxena2019data}. We only cover a subset of these approaches here due to lack of space and refer the interested reader to a recent survey on learning with noisy labels~\citep{han2021survey}.

\vspace{-2mm}
\section{Experiments}
\vspace{-2mm}
\label{sec:exp}

\begin{table*}[!ht]
\vspace{-4mm}
\caption{{\small \bf Test accuracy  on CIFAR-10 and CIFAR-100 with symmetric label noise using ResNet-18.} Noise rate is varied in the range $\{0.2, 0.4, 0.6, 0.8\}$. %
Blank cells for CICW and CICW-M denote that we do not get improvement over CIW with a $\gamma>0$ (Obj. \eqref{prob:class_w}).
Methods marked $^{\boldsymbol{\ddagger}}$ need to either maintain scalar/vector weights over full train set, or fit a model on loss values of full train set (extra forward pass for entire train set per epoch).}
\label{tab:cifar}
\centering
\small
\resizebox{\textwidth}{!}{
\begin{tabular}{c|ccccc}
\toprule
 \multirow{2}{*}{Methods} &  \multicolumn{5}{c}{Noise Rate ({\bf CIFAR-10})}\\
&  Clean & 0.2 & 0.4 & 0.6 & 0.8 \\ \hline \hline
  CE & 92.44 $\pm$ 0.17 & 84.03 $\pm$ 0.38 & 78.56 $\pm$ 0.98 & 67.64 $\pm$ 1.12 & 36.86 $\pm$ 1.96 \\
  Bi-tempered \citep{bitempered}& 92.55 $\pm$ 0.10 & 89.74 $\pm$ 0.22 & 83.87 $\pm$ 0.37 & 72.49 $\pm$ 0.77 & 40.41 $\pm$ 1.09 \\
  APNL \citep{ma2020normalized} & 90.93 $\pm$ 0.16 & 88.63 $\pm$ 0.35 & 84.97 $\pm$ 0.35 & 77.47 $\pm$ 0.46 & 42.71 $\pm$ 1.47 \\ \midrule
   EG$^{\boldsymbol{\ddagger}}$ \citep{bar2021multiplicative,majidi2021exponentiated} & 92.29 $\pm$ 0.12 & 90.22 $\pm$ 0.16 & 86.48 $\pm$ 0.33 & 80.03 $\pm$ 0.74 & 45.38 $\pm$ 2.24 \\
   ELR$^{\boldsymbol{\ddagger}}$ \citep{liu2020early} & -- & 91.15 $\pm$ 0.05 & 88.38 $\pm$ 0.44& 78.94 $\pm$ 0.47& 40.15 $\pm$ 3.28\\
 \midrule
 CIW ($\alpha=1$) & 92.14 $\pm$ 0.20& 88.74 $\pm$ 0.26 & 85.13 $\pm$ 0.28 & 78.27 $\pm$ 0.41 & 43.07 $\pm$ 0.34 \\
\midrule
 CICW ($\alpha=1$) & -- & 89.49 $\pm$ 0.14 & 86.45 $\pm$ 0.34 & 78.68 $\pm$ 0.37 & 44.03 $\pm$ 1.87 \\
 \midrule
Mixup \citep{zhang2017mixup} & 93.66 $\pm$ 0.20 & 89.48 $\pm$ 0.24 & 85.00 $\pm$ 0.46 & 75.12 $\pm$ 0.67 & 44.44 $\pm$ 1.96 \\
Dyn-Mixup$^{\boldsymbol{\ddagger}}$ \citep{arazo2019unsupervised} & 92.52 $\pm$ 0.22 & 89.69 $\pm$ 0.15 & 87.30 $\pm$ 0.33 & 78.04 $\pm$ 0.59 & 46.04 $\pm$ 0.88 \\ \midrule 
CICW-M ($\alpha=1$) & -- & {\bf 91.34 $\pm$ 0.16} & {\bf 89.95 $\pm$ 0.15} & {\bf 84.53 $\pm$ 0.27} & {\bf 58.74 $\pm$ 0.73} \\
\midrule
\midrule
&  \multicolumn{5}{c}{Noise Rate ({\bf CIFAR-100})}\\
&  Clean & 0.2 & 0.4 & 0.6 & 0.8 \\ \hline \hline
  CE & 70.52 $\pm$ 0.32 & 56.09 $\pm$ 0.39 & 43.34 $\pm$ 1.17 & 29.41 $\pm$ 1.00 & 12.47 $\pm$ 1.09 \\
  Bi-tempered \citep{bitempered} & 72.01 $\pm$ 0.26 & 67.51 $\pm$ 0.29 & 60.60 $\pm$ 0.71 & 47.25 $\pm$ 0.74 & 20.89 $\pm$ 0.64 \\
  APNL \citep{ma2020normalized}& 67.62 $\pm$  0.45 & 65.66 $\pm$ 0.37 & 60.25 $\pm$ 0.43 & 47.28 $\pm$ 0.97 & 20.12 $\pm$ 0.93\\ \midrule
   EG$^{\boldsymbol{\ddagger}}$ \citep{bar2021multiplicative,majidi2021exponentiated} & 70.38 $\pm$ 0.19 & 67.15 $\pm$ 0.35 & 61.59 $\pm$ 0.32 & 50.58 $\pm$ 0.58 & 25.76 $\pm$ 0.61 \\
   ELR$^{\boldsymbol{\ddagger}}$ \citep{liu2020early} & -- & 65.73 $\pm$ 0.24& 57.37 $\pm$ 0.48& 40.64 $\pm$0.51& 12.87 $\pm$ 0.57 \\
 \midrule
 CIW ($\alpha=1$) & 70.08 $\pm$ 0.44 & 65.89 $\pm$ 0.30        & 60.54 $\pm$ 0.45 & 50.53 $\pm$ 0.37 & 25.00 $\pm$ 0.80 \\
\midrule
 CICW ($\alpha=1$) & -- & -- & 60.85 $\pm$ 0.44 & 50.57 $\pm$ 0.46 & 25.36 $\pm$ 0.92 \\
 \midrule
Mixup \citep{zhang2017mixup} & 72.63 $\pm$ 0.22 & 65.48 $\pm$ 0.27 & 58.61 $\pm$ 0.32  & 45.46 $\pm$ 0.94 & 21.86 $\pm$ 0.65 \\
Dyn-Mixup$^{\boldsymbol{\ddagger}}$ \citep{arazo2019unsupervised} & 71.28 $\pm$ 0.33 & 67.74 $\pm$ 0.24 & 61.14 $\pm$ 0.48 & 49.37 $\pm$ 0.48 & 26.48 $\pm$ 1.19 \\ \midrule 
CICW-M ($\alpha=1$) & -- & {\bf 68.23 $\pm$ 0.23} & {\bf 63.64 $\pm$ 0.45} & {\bf 54.46 $\pm$ 0.72} & {\bf 28.94 $\pm$ 0.74} \\
\bottomrule
\end{tabular}
}
\vspace{-0.3cm}
\end{table*}

We evaluate our method on three standard benchmark datasets: CIFAR-10, CIFAR-100, and Clothing1M \citep{xiao2015learning}. Following earlier works, we experiment with  synthetic noise for the CIFAR datasets while Clothing1M naturally has noisy labels. CIFAR-10 and CIFAR-100 have 32x32 RGB images with 10 and 100 classes, respectively. We apply standard data augmentation of padding to 36$\times$36 followed by crop to 32$\times$32, and random horizontal flipping for both CIFAR datasets. Clothing1M has 256$\times$256 RGB images which we center crop to size 224$\times$224, and has 14 classes. Next, we describe our experimental setup in more detail. 

{\bf Architecture.~} We use a PreAct ResNet-18 \citep{he2016identity} for CIFAR-10 and CIFAR-100. Following earlier work, we use a ImageNet pretrained ResNet50v2 \citep{he2016identity} for Clothing1M, replacing its last dense layer with another dense layer having 14 outputs, and finetune it on Clothing1M.

{\bf Baselines.~} Apart from the standard cross-entropy {\bf (CE)} loss optimization, we empirically compare our approach with several recently proposed methods for addressing label noise:  {\bf (i) Bi-tempered loss} \citep{bitempered} is based upon tempered- exponential and logarithm, and has been shown to be robust to label noise. {\bf (ii) APNL} or active-passive normalized losses \citep{ma2020normalized} combine an \emph{active} normalized loss with a \emph{passive} normalized loss and are shown to be quite effective for label noise \citep{ma2020normalized}. 
{\bf (iii) EG} or exponentiated-gradient \citep{bar2021multiplicative,majidi2021exponentiated} was also recently proposed for tackling the problem of label noise and showed strong empirical performance. However, it needs to maintain a record of importance weights for the full training set that need to be updated throughout the training, which results in increased overhead. {\bf (iv) Mixup} \citep{zhang2017mixup} uses a random convex combination of training examples to regularize the loss landscape and has been shown earlier to help with label noise. {\bf (v)} Dynamic Mixup with bootstrapping {\bf (Dyn-Mixup)} \citep{arazo2019unsupervised} makes use of the observation that noisy examples have higher losses than the clean examples (particularly early on in training) and fits a two-component beta-mixture distribution on the training losses at every epoch using EM \citep{dempster1977maximum}. The posterior probabilities for each example are used in mixing examples with Mixup. In other words, in Dyn-Mixup, the inputs are mixed using the importance weights (normalized by their sum) while the labels are mixed randomly (similar to Mixup). Similar to EG \citep{bar2021multiplicative,majidi2021exponentiated}, this approach also results in increased overhead due to the need to fit the beta-mixture on losses for full training set. 
{\bf (vi)} Early-learning regularization or {\bf ELR} \cite{liu2020early} adds a regularizer to the objective that encourages alignment of current model predictions to weighted average of past predictions. It needs to maintain a $k$ dimensional vector for each training example. 

{\bf Proposed methods.~} As we propose a family of optimization problems that can lead to several variants depending on the divergence used, here we summarize the variants we experiment with in this work: {\bf (i) CIW}: %
We mainly experiment with $\alpha$-divergences for $D(u,w)$, restricting ourselves to $\alpha=0$ (Reverse-\kl~divergence), $\alpha=0.5$ and $\alpha=1$ (\kl-divergence). The results for $\alpha=0$ and $0.5$ are reported in the Appendix. {\bf (ii) CICW}: %
We fix the divergence $D_1$ for the instance weights to \kl-divergence ($\alpha=1$) and experiment with both total-variation and $\ell_2$- distances for $D_2$. {\bf (iii) CICW-M} (CICW with Mixup): Apart from reweighting both instances and classes, we also use instance weights with Mixup \citep{zhang2017mixup} as described in Sec.~\ref{subsec:mixup}. This is similar in spirit to dynamic mixup (Dyn-Mixup) \citep{arazo2019unsupervised} which uses posterior probabilities of beta-mixture distribution with Mixup. We again fix $D_1$ to be \kl-divergence, and $D_2$ to be \kl-divergence for CIFAR-10 and $\ell_1$ for CIFAR-100. %
We experiment with the two possibilities of using importance weights with Mixup as discussed in Sec.~\ref{subsec:mixup} (taking it as a binary hyperparameter).

\begin{table*}[!ht]
\vspace{-4mm}
\caption{{\small \bf Test accuracy  on CIFAR-10 and CIFAR-100 with asymmetric label noise using ResNet-18.} Noise rate is varied in the range $\{0.1, 0.2, 0.3, 0.4\}$. %
Methods marked with $^{\boldsymbol{\ddagger}}$ need to either maintain scalar/vector weights over full train set, or fit a model on loss values of full train set (extra forward pass for entire train set per epoch).
}
\label{tab:cifar_asym}
\centering
\small
\resizebox{\textwidth}{!}{
\begin{tabular}{c|ccccc}
\toprule
 \multirow{2}{*}{Methods} &  \multicolumn{5}{c}{Noise Rate ({\bf CIFAR-10})}\\
&  Clean & 0.1 & 0.2 & 0.3 & 0.4 \\ \hline \hline
  CE & 92.44 $\pm$ 0.17 & 89.53 $\pm$ 0.13 & 86.53 $\pm$ 0.65 & 83.67 $\pm$ 1.26 & 76.67 $\pm$ 1.30 \\
  Bi-tempered \citep{bitempered}& 92.55 $\pm$ 0.10 & 91.24 $\pm$ 0.16 & 89.53 $\pm$ 0.29 & 86.46 $\pm$ 0.66 & 81.43 $\pm$ 1.32 \\
  APNL \citep{ma2020normalized} & 90.93 $\pm$ 0.16 & 89.59 $\pm$ 0.36 & 88.22 $\pm$ 0.23 & 85.05 $\pm$ 0.45 & 80.45 $\pm$ 0.17 \\ \midrule
   EG$^{\boldsymbol{\ddagger}}$ \citep{bar2021multiplicative,majidi2021exponentiated} & 92.29 $\pm$ 0.12 & 91.41 $\pm$ 0.12 & 90.07 $\pm$ 0.14 & 88.75 $\pm$ 0.15 & 85.55 $\pm$ 0.24 \\
 \midrule
 CIW ($\alpha=1$) & 92.14 $\pm$ 0.20 & 90.40 $\pm$ 0.33 & 88.77 $\pm$ 0.23 & 88.70 $\pm$ 0.28 & 86.56 $\pm$ 0.17\\
\midrule
 CICW ($\alpha=1$) & -- & 90.55 $\pm$ 0.14 & 89.01 $\pm$ 0.25 & -- & 86.34 $\pm$ 0.44 \\
 \midrule
 ELR$^{\boldsymbol{\ddagger}}$ \citep{liu2020early} & -- & 92.70 $\pm$ 0.17 & 91.90 $\pm$ 0.24 & 90.93 $\pm$ 0.13 & 87.40 $\pm$ 0.58\\
 CIW-ELR$^{\boldsymbol{\ddagger}}$ & -- & 92.62 $\pm$ 0.19 & 91.92 $\pm$ 0.17 & {\bf 91.31 $\pm$ 0.20} & {\bf 89.80 $\pm$ 0.16}\\
 \midrule
Mixup \citep{zhang2017mixup} & 93.66 $\pm$ 0.20 & 91.78 $\pm$ 0.71 & 91.20 $\pm$ 0.49 & 90.81 $\pm$ 0.29 & 88.05 $\pm$ 0.87 \\
Dyn-Mixup$^{\boldsymbol{\ddagger}}$ \citep{arazo2019unsupervised} & 92.52 $\pm$ 0.22 & 91.08 $\pm$ 0.22 & 90.48 $\pm$ 0.27 & 90.24 $\pm$ 0.20 & 87.34 $\pm$ 0.32  \\ \midrule 
Dyn-CICW-M ($\alpha=1$)& -- & {\bf 93.06 $\pm$ 0.09} & {\bf 92.44 $\pm$ 0.48} & 90.91 $\pm$ 0.28 & 86.41 $\pm$ 0.51 \\
\midrule
\midrule
&  \multicolumn{5}{c}{Noise Rate ({\bf CIFAR-100})}\\
&  Clean & 0.1 & 0.2 & 0.3 & 0.4 \\ \hline \hline
  CE & 70.52 $\pm$ 0.32 & 64.25 $\pm$ 0.17 & 58.55 $\pm$ 0.19  & 52.35 $\pm$ 0.64 & 45.81 $\pm$ 0.49 \\
  Bi-tempered \citep{bitempered}& 72.01 $\pm$ 0.26 & 68.87 $\pm$ 0.33 & 66.40 $\pm$ 0.41 & 63.75 $\pm$ 0.21 & 60.19 $\pm$ 0.51 \\
  APNL \citep{ma2020normalized} & 67.62 $\pm$  0.45 & 66.93 $\pm$ 0.22 & 63.78 $\pm$ 0.48 & 59.75 $\pm$ 0.33 & 56.14 $\pm$ 0.29 \\ \midrule
   EG$^{\boldsymbol{\ddagger}}$ \citep{bar2021multiplicative,majidi2021exponentiated} & 70.38 $\pm$ 0.19 & 68.50 $\pm$ 0.29 & 65.88 $\pm$ 0.28 & 63.94 $\pm$ 0.39 & 61.53 $\pm$ 0.28 \\
 \midrule
 CIW ($\alpha=1$) & 70.08 $\pm$ 0.44 & 67.42 $\pm$ 0.20 & 66.54 $\pm$ 0.28 & 64.68 $\pm$ 0.17 & 62.40 $\pm$ 0.21 \\
\midrule
 CICW ($\alpha=1$) & -- & 67.97 $\pm$ 0.26 & 66.63 $\pm$ 0.48 & 64.55 $\pm$ 0.29 & 62.53 $\pm$ 0.37 \\
 \midrule
 ELR$^{\boldsymbol{\ddagger}}$ \citep{liu2020early} & -- & 69.40 $\pm$ 0.21 & 67.69 $\pm$ 0.24 & 66.03 $\pm$ 0.34 & 64.41 $\pm$ 0.55\\
 CIW-ELR$^{\boldsymbol{\ddagger}}$ & -- & 69.63 $\pm$ 0.27 & 68.31 $\pm$ 0.21 & 66.77 $\pm$ 0.32 & 65.18 $\pm$ 0.31\\
 \midrule
Mixup \citep{zhang2017mixup} & 72.63 $\pm$ 0.22  & 69.06 $\pm$ 0.25 & 66.37 $\pm$ 0.33 & 62.78 $\pm$ 0.42 & 59.76 $\pm$ 0.30 \\
Dyn-Mixup$^{\boldsymbol{\ddagger}}$ \citep{arazo2019unsupervised} & 71.28 $\pm$ 0.33 & 69.44 $\pm$ 0.27 & 67.95 $\pm$ 0.23 & 65.58 $\pm$ 0.52 & 62.27 $\pm$ 0.37 \\ \midrule 
Dyn-CICW-M ($\alpha=1$)& -- & {\bf 71.64 $\pm$ 0.24} & {\bf 70.08 $\pm$ 0.40} & {\bf 68.42 $\pm$ 0.19} & {\bf 65.93 $\pm$ 0.27} \\
\bottomrule
\end{tabular}
}
\vspace{-0.3cm}
\end{table*}

{\bf Hyperparameter search.~} As CIFAR-10 and CIFAR-100 do not have a separate validation set, we randomly split the training examples of each of these into two subsets: 90\% are used for training while the rest 10\% are used for validation. This leaves 5000 examples in the validation sets of both datasets. %
We also add same amount of noise to the validation set labels as in the train set, \ie, we \emph{do not} assume a clean validation set for the CIFAR datasets. This %
differs from some earlier works that do not use a noisy validation set for hyperparameter search \cite{dynmix-code-acc-reporting,napl-code-acc-reporting} and directly report the best accuracy on CIFAR test sets across all hyperparameters. %
Our assumption of availability of a noisy validation set is realistic.  %
We perform hyperparameter search for all methods using the validation set. More details on hyperparameters are provided in the Appendix. The performance of the proposed methods is reasonably robust to hyperparameter variation and we provide sensitivity plot for CIW in the Appendix. 
\remove{
{\bf Our methods have following hyperparameters}: (i) CIW has one hyperparameter, $\lambda$ for \kl-divergence, (Eq.~\eqref{eq:w_kldiv}), or hyperparameter $\mu$ for reverse-\kl~divergence (Eq.~\eqref{eq:w_revkl}) and $\alpha$-divergence for $\alpha=0.5$ (Eq.~\eqref{eq:w_alphadiv_raw}). (ii) CICW has two additional hyperparameters: the choice of divergence $D_2$ (total-variation or $\ell_2$), and parameter $\gamma$ in the objective \eqref{prob:class_w}. (iii) CICW-M has two additional hyperparameters over CICW: the choice of how to use importance weights in Mixup (IW-Mix or SIW-Mix in Sec.~\ref{subsec:mixup}), and the choice of how to use mixed up examples for training (Mixup-base or Mixup-reweight in Sec.~\ref{subsec:mixup}). For CIFAR-10 we fix $D_2$ to \kl-divergence for CICW-M, and for CIFAR-100 we search it over total-variation or $\ell_2$ distance. We use a \emph{burn-in} period of 4000 minibatch iterations for CIFAR-10 and CIFAR-100 where we train the network using standard cross-entropy loss before activating the instance and class reweighting. We use a burn-in period of 2000 iterations for Clothing1M. }
{\bf Optimization.~} %
We use SGD optimizer with a momentum of $0.9$ with Nesterov acceleration. More details on optimization are provided in the Appendix. 

\remove{We train the ResNet-18 model for both CIFAR datasets using SGD optimizer with a momentum of $0.9$ with Nesterov acceleration. We use initial learning rate of 0.1 and a piecewise constant learning rate schedule of $(10^{-2}, 10^{-3}, 10^{-4})$ at $(30,80,110)$ epochs, respectively. We use a batch size of $128$ and train the model for $140$ epochs. For the ResNet50v2 model for Clothing1M, we again use a SGD optimizer with a momentum of $0.9$ with Nesterov acceleration and additionally use a weight decay. We tune the initial learning rate and weight decay parameter for the cross-entropy baseline using Clothing1M validation set and fix them to these values for all other methods ($0.005$ for learning rate and $10^{-5}$ for weight decay). We use a batch size of 64 and finetune it for total 8 epochs, while  reducing the learning rate to $0.0005$ after 5 epochs. We implement all methods using TensorFlow \citep{tensorflow2015-whitepaper} (Apache 2.0 license), and train all models on Nvidia V100 GPUs.} 

{\bf Results.~} The clean test accuracy results for all the methods for CIFAR-10 and CIFAR-100 are shown in Tables \ref{tab:cifar} and \ref{tab:cifar_asym}. %
Table \ref{tab:cifar} shows the results for symmetric noise
for noise rates $\eta\in\{0.2, 0.4, 0.6, 0.8\}$, which means the label of each training and validation example is \emph{flipped}\footnote{This is different from some earlier works such as \cite{arazo2019unsupervised} where noisy label is \emph{randomly sampled} from all possible labels which results in effectively a lower noise rate.} to the other labels with probability $\eta$. As expected, all methods designed for noisy labels greatly improve over the CE baseline, more so for higher noise levels. Among the baselines, EG \citep{bar2021multiplicative,majidi2021exponentiated} and Dynamic-mixup \citep{arazo2019unsupervised} are the most competitive and come close to our methods CIW and CICW. However, both these methods need to maintain a distribution of weights over the full training set across the training iterations which creates overhead and may not fit into the standard pipelines for training production ML systems. It is notable that despite the simplicity our methods are able to match the performance of these methods.  Furthermore, Dyn-Mixup \citep{arazo2019unsupervised} also uses Mixup while CIW and CICW do not. The proposed CIW and CICW outperform other baselines, including Bi-tempered loss \citep{bitempered}, APNL \citep{ma2020normalized} and Mixup \citep{zhang2017mixup}, particularly for high noise cases. Finally, we observe that the proposed CICW-M %
outperforms all baselines by a significant margin, indicating that CIW weights can be quite effective when used with Mixup.  

Table \ref{tab:cifar_asym} shows the results for asymmetric noise with noise rates $\eta\in\{0.1, 0.2, 0.3, 0.4\}$. We use same asymmetric noise as used in earlier works \cite{ma2020normalized}. We provide the details in the Appendix for completeness. ELR \cite{liu2020early} is the best performing baseline in this case but it needs to maintain a $K$-dimensional vector for every training example in the train set. ELR is also complementary to our method and its regularizer can be added to our objective for further improvements (the CIW-ELR row in the table). Our dynamic mixup variant with CIW weights as described in Sec.~\ref{subsec:mixup}, Dyn-CICW-M, yields significant improvement over all baselines including the ones that use Mixup \citep{zhang2017mixup,arazo2019unsupervised}, particularly for CIFAR-100. CICW-M did not improve over CICW for asymmetric noise so we do not report its results. We believe this is due to the fact that using CIW weights for Mixup in the case of asymmetric noise  biases the mixed up images towards certain mixture proportions.

We also experiment with Clothing1M dataset. Our CE baseline gives stronger accuracy of 70.32$\pm$0.28 compared to what has been earlier reported (due to our tuning of the learning rate and weight decay parameter). We find the all the methods remain in the same range and are unable to claim any significant improvement over this baseline. EG \citep{bar2021multiplicative} gives 70.52$\pm$0.50, APNL \citep{ma2020normalized}  69.35$\pm$0.57, Vanilla Mixup 70.48$\pm$0.21, and Dyn-Mixup \citep{arazo2019unsupervised} gives 
70.33$\pm$0.27. The proposed CIW 
($\alpha=0.5$) gives 70.52$\pm$0.46, and CICW ($\alpha=1$) gives 70.84$\pm$0.28. 
While we stil get an improvement with the proposed CICW, it is much less pronounced than our results on CIFAR-10 and CIFAR-100. We also observed that Mixup with CIW weights did not yield much gains.

\vspace{-4mm}
\section{Conclusion}
\label{sec:discuss}
\vspace{-3mm}
We proposed a class of constrained optimization problems for tackling label noise that yield simple closed form updates for reweighting the training instances and class labels. We also proposed ways for using the instance weights with Mixup that results in further significant performance gains over instance and class reweighting. 
Our method operates solely at the level of minibatches which avoids the extra overhead of maintaining dataset level weights as in earlier methods \citep{bar2021multiplicative,majidi2021exponentiated,arazo2019unsupervised}.
As part of the limitation, it still remains to be seen how well the method works on other realistic noisy label settings that are encountered in practice.
We believe that studying the interaction of our framework with label smoothing is an interesting direction for future work that can result in a loss adaptive version of label smoothing. \\

\noindent {\bf Acknowledgement.~} We are thankful to Kevin Murphy for providing several helpful comments on the manuscript.

\bibliography{ml}
\bibliographystyle{plain}

\onecolumn
\appendix
\begin{center}
{\large \bf Appendix} \\ 
\end{center}
Here we provide more details, in particular

{\bf Appendix
\ref{sec:upper_b}}: Proof of Theorem \ref{thm:upper_b} showing that our minibatch objective optimizes an upper bound on the population objective in expectation.\\
{\bf Appendix \ref{app:f_div}}: Proof of Theorem \ref{thm:w_fdiv} showing the weight updates for $f$-divergence constrained objective \\
{\bf Appendix \ref{app:bregman}}: Constrained instance reweighting updates for Bregman divergence constraint. \\
{\bf Appendix
\ref{app:cicw_proofs}}: Solution to constrained class reweighting objective for total variation distance, $\ell_2$ distance, $\ell_\infty$ distance, and \kl-divergence. \\
{\bf Appendix \ref{app:twomoons}}: Details on experimental setup for decision boundary visualization on the two-moons dataset. \\
{\bf Appendix \ref{app:algorithm}}: Algorithmic sketch of the proposed method. \\
{\bf Appendix \ref{app:opt_details}}: Details on optimization of the model. \\
{\bf Appendix \ref{app:label_noise}}: Details about symmetric and asymmetric noise. \\
{\bf Appendix \ref{app:exp_alpha}}: Empirical results for other $\alpha$-divergences. \\
{\bf Appendix \ref{app:hyperparams}}: More details on hyperparameter selection and plots for hyperparameter sensitivity. 

\section{Proof of Theorem \ref{thm:upper_b}: upper bound on the population objective}
\label{sec:upper_b}
Our proof relies on the techniques used in \citep{shapiro2017distributionally,levy2020large} for the analysis of distributionally robust objectives. 
We recall our proposed population objective:
\begin{align}
\inf_\theta \inf_Q \expect_{x\sim Q} L(x,\theta),\, \text{s.t. } D_f(Q,P)\leq\delta,
\end{align}
\noindent where $P$ is the data distribution and $D_f$ is the $f$-divergence with $D_f(Q,P)=\int f(dQ/dP)dP$. We define 
\begin{align}
L(P;\theta):=\inf_{Q:D_f(Q,P)\leq\delta} \expect_{Q} L(x,\theta)
\label{appeq:pop_obj}
\end{align}
and 
\begin{align}
L(S_{1:n};\theta) := L(\hat{P}_n;\theta) = \min_{w:w\geq 0, \lVert w\rVert_1=1, \frac{1}{n}\sum_{i=1}^n  f(nw_i)\leq \delta} \sum_i w_i L(x_i,\theta)\, ,    
\label{appeq:sample_obj}
\end{align}
\noindent where $\hat{P}_n$ is the uniform distribution over samples $S_{1:n}=\{x_i\}_{i=1}^n$ (sampled from data distribution $P$). We can rewrite \eqref{appeq:pop_obj} using the inverse cdf of $L(x,\theta)$. Let $F^{-1}$ be the inverse cdf of $L(x,\theta)$ under $P$ which implies that the distribution of $L(\theta,x)$ for $x\sim P$ is equal to the distribution of $F^{-1}(u)$ for $u\sim U$ with $U=\text{Unif}(0,1)$. Hence we have
\begin{align}
L(P;\theta) = \inf_{Q':D_f(Q',U)\leq\delta} \expect_{u\sim Q'} F^{-1}(u) = \inf_{r\in R} \int_0^1 r(u)F^{-1}(u)\, du\, ,
\label{appeq:obj_cdf}
\end{align}
\noindent where $r(u)=\frac{dQ'}{dU}(u)$ and the set $R = \{r:[0,1]\to \reals_+ | \int_0^1 r(u)\,du=1\, \text{and } \int_0^1 f(r(u))\,du\leq \delta\}$. 
The problem \eqref{appeq:obj_cdf} is linear objective on a convex set and the problem is strictly feasible, hence strong duality holds. The dual of the problem \eqref{appeq:obj_cdf} is given by $L(P;\theta)=\sup_{\mu,\lambda\geq 0} G(P;\theta,\mu,\lambda)$, where
\begin{align}
G(P;\theta,\mu,\lambda) := \int_0^1 \inf_{r\in\reals_+} [rF^{-1}(u) + \mu(r-1) + \lambda(f(r)-\delta)]\, du\, .
\end{align}
Note that we interchanged the order of $\inf$ and the integral following \cite{shapiro2017distributionally,rockafellar2009variational}. 
Writing $\inf_{r\in\reals_+}(rF^{-1}(u) + \mu r + \lambda f(r)) = \sup_{r\in\reals_+} (r(-F^{-1}(u)-\mu) - \lambda f(r)) := (\lambda f)^*(r(-F^{-1}(u)-\mu))$ (the convex conjugate of function $r\mapsto\lambda f(r))$, we get
\begin{align}
\begin{split}
G(P;\theta,\mu,\lambda) & = \int_0^1 (\lambda f)^* (-\mu-F^{-1}(u))\, du - \mu - \lambda\delta \\
&= \expect_{x\sim P} (\lambda f)^*(-\mu - L(\theta,x)) -\mu - \lambda\delta \, .
\end{split}
\end{align}
The second equality above is due to the fact that the distribution of $L(\theta,x)$ for $x\sim P$ is same as the distribution of $F^{-1}(u)$ for $u\sim U$ with $U=\text{Unif}(0,1)$.
Defining $P_{\times n}$ as the distribution over sets of elements (size $n$) such that each element of set is iid sampled from $P$, we have
\begin{align}
\begin{split}
L(P;\theta) = \sup_{\mu,\lambda\geq 0} G(P;\theta,\mu,\lambda) &= \sup_{\mu,\lambda\geq 0} \expect_{x\sim P}  [(\lambda f)^* (-\mu-L(\theta,x))] - \mu - \lambda\delta \\
&= \sup_{\mu,\lambda\geq 0} \expect_{S_{1:n}\sim P_{\times n}} \frac{1}{n}\sum_{x\in S_{1:n}} [(\lambda f)^* (-\mu-L(\theta,x))] - \mu - \lambda\delta \\
& \leq \expect_{S_{1:n}\sim P_{\times n}} \sup_{\mu,\lambda\geq 0} \frac{1}{n}\sum_{x\in S_{1:n}} [(\lambda f)^* (-\mu-L(\theta,x))] - \mu - \lambda\delta \\
&= \expect_{S_{1:n}\sim P_{\times n}} L(S_{1:n};\theta)\, .
\end{split}
\end{align}
Hence, our approach that optimizes for the instance weights per minibatch minimizes an upper bound (in expectation) on the population version of the objective. This is a reassurance that we are \emph{not} minimizing a lower bound on the population objective. 

\section{Proof of Theorem \ref{thm:w_fdiv}: weight update for $f$-divergence constrained problem}
Recall the finite-sample version of our problem 
\begin{align}
L(\hat{P}_n;\theta) := \min_{w:w\geq 0, \lVert w\rVert_1=1, D(w,u)\leq \delta} \sum_i w_i L(x_i,\theta)\, .    
\end{align}
\label{app:f_div}
Forming the Lagrangian for the problem, we get
\begin{align}
\sum_i w_i L(x_i,\theta) + \lambda (D(w,u)-\delta) + \mu(\sum_i w_i - 1) -\sum \nu_i w_i\,,
\label{eq:gen_lag}
\end{align}
\noindent where $\lambda\geq 0, \nu_i\geq 0, \mu$ are the Lagrange multipliers. 
The dual function (for a fixed $\theta$) is given by
\begin{align}
h(\lambda,\mu,\nu) = \min_w \sum_i w_i L(x_i,\theta) + \lambda (D(w,u)-\delta)+ 
\mu(\sum_i w_i - 1) - \sum_i \nu_i w_i\, .
\label{eq:gen_dual}
\end{align}
Optimizing over $w$, the first order condition for optimality is
\begin{align}
\begin{split}
& L(x_i,\theta) + \lambda  f'\left(\frac{w_i}{u_i}\right) + \mu - \nu = 0\\
\Longrightarrow\, & w_i = u_i f'^{-1}\left(\frac{-L(x_i,\theta)-\mu+\nu_i}{\lambda}\right)= \frac{1}{n} f'^{-1}\left(\frac{-L(x_i,\theta)-\mu+\nu_i}{\lambda}\right)\, .
\end{split}
\label{eq:w_fdiv}
\end{align}
The parameters $\mu, \lambda$ and $\nu_i$ are such that the constraints are satisfied. 

\section{Constrained instance reweighting using Bregman divergence}
\label{app:bregman}
The Bregman divergence using a convex function $F$ is defined as
\[
D(u, w) = F(u) - F(w) - \nabla F(w)\cdot (u - w)\, .
\]
For convenience, we use $f \coloneqq \nabla F$. We again work with inverse Bregman divergence $D(w,u)$ to obtain  closed form updates. The Lagrangian dual is given by  
\begin{align}
h(\lambda,\mu,\nu) = \min_w \sum_i w_i L(x_i,\theta) + \lambda (D(w,u)-\delta)+ \mu(\sum_i w_i - 1) - \sum_i \nu_i w_i\, .
\label{eq:app_gen_dual}
\end{align}
First order optimality condition for $w$ (for a fixed $\theta$) is given by
\begin{align}
L(\theta) + \lambda (f(w) - f(u)) + \mu\,1 - \nu = 0 \Longrightarrow w = f^{-1}\left(f(u) - \frac{L(\theta) + \mu1 - \nu}{\lambda}\right)\,,
\end{align}
where $L(\theta)$, and $\nu$ are the vectors of losses and Lagrange multipliers, respectively, and $\mu1$ is a vector with each entry equal to $\mu$. The Lagrange multipliers $\lambda$, $\mu$ and $\nu_i$ are such that the constraints are satisfied. 

\section{Constrained class reweighting}
\label{app:cicw_proofs}
We consider the following class reweighting problem:
\begin{align}
\min_v \sum_j v_{j} L_j(x,\theta),\quad \text{s.t. }  D_2(e ,v) \leq \gamma, \sum_j v_{j}=1, v_{j} \geq 0\,,
\label{prob:app_class}
\end{align}
where $L_j(x,\theta)$ is the loss of example $x$ assuming the true label is $j$, \ie, $L_j(x,\theta)\coloneqq L(x,j,\theta)$, and $e$ is a one-hot vector with $1$ at the index of true label, \ie, $e_y=1$. We provide solutions to this problem when $D_2$ is total variation distance, $\ell_2$ distance, $\ell_\infty$ distance and reverse $f$-divergence (such as reverse \kl~divergence). 

\subsection{Proof of Theorem \ref{thm:v_totalvar}: Class reweighting with total variation distance}
\label{app:l1_dist}
Taking $D_2$ to be the total variation distance will result in a linear program in $v$ with solution lying on a vertex. We can rewrite the optimization problem \eqref{prob:app_class} in $v$ as 
\begin{align}
\min_v \sum_j v_{j} L_j(x,\theta),\quad \text{s.t. }  \lVert e - v\rVert_1 \leq \gamma, \sum_j v_{j}=1, v_{j} \geq 0\,,
\label{prob:app_class_tv}
\end{align}
where $e$ is a one-hot vector with $e_y=1$ and $e_j=0\, \forall j\neq y$.

We can write $\lVert e - v\rVert_1=(1-v_{y}) + \sum_{j\neq y} v_{j}$, where $y$ is the label of example $x$. Since $\sum_j v_{j}=1$, the constraint $\lVert e - v\rVert_1\leq \gamma$ transforms into $v_y\geq 1-\gamma/2$ and $\sum_{j\neq y} v_j \leq \gamma/2$. Substituting it in \eqref{prob:app_class} yields
\begin{align}
\min_v \sum_{j} v_{j} L_j(x,\theta),\quad \text{s.t. }  v_y\geq 1-\gamma/2, \sum_{j\neq y}v_j \leq \gamma/2, \sum_j v_j = 1, v_{j} \geq 0\, .
\label{prob:app_class_tv2}
\end{align}
If $y=\argmin_j L_j(x,\theta)$, the solution of the linear program in \eqref{prob:app_class_tv2} will be the one-hot vector with $1$ at the position $y$. For the case $\hat{y}=\argmin_{j} L_j(x,\theta)\neq y$, the solution will be a two-hot vector with $1-\gamma/2$ at the position $y$ and $\gamma/2$ at the position $\hat{y}$. Plugging the optimal values for $v$ back into the objective \eqref{prob:class_w} reduces it to
\begin{equation}
\begin{split}
\min_{w,\theta} \sum_i w_i \left[(1-\gamma/2) L_{y_i}(x_i,\theta) + \gamma/2 L_{\hat{y_i}}(x_i,\theta)\right] \text{ s.t. } D_1(u,w)\leq \delta, \sum_i w_i = 1, w_i \geq 0\,.
\end{split}
\end{equation}

\subsection{Class reweighting with $\ell_2$-distance}
\label{app:l2_dist}
We now take $D_2$ to be the squared $\ell_2$-distance  and consider the following problem:
\begin{align}
\min_v \sum_j v_{j} L_j(x,\theta),\quad \text{s.t. }  \lVert e - v\rVert_2^2 \leq \gamma, \sum_j v_{j}=1, v_{j} \geq 0\,.
\label{prob:app_class_l2}
\vspace{-3mm}
\end{align}
Lagrangian of the above problem is given by $\mcal{L}(\theta,\lambda,\mu,\nu)=\sum_j v_{j} L_j(x,\theta) + \lambda/2 (\lVert v - e\rVert^2 - \gamma) + \mu(\sum_j v_j -1) - \sum_j \nu_j v_j$, 
where $\lambda\geq 0$ and $\nu_j\geq 0$. Minimizing it w.r.t. $\theta$ gives
\begin{align}
v_j = \frac{-1}{\lambda}(L_j(x,\theta)-\nu_j+\mu) + e_j\, .
\label{eq:app_v_l2_cls}
\end{align}
Since $\sum_j v_j=1$, we get $\mu=-\frac{1}{k}\sum_j(L_j(x,\theta)-\nu_j)$, where $k$ is the number of classes, $e_y=1$ and $e_j=0$ for $j\neq y$. 
When $y=\argmin_j L_j(x,\theta)$, the solution of \eqref{eq:app_v_l2_cls} will be the one-hot vector with $1$ at the position $y$ and the constraint $\lVert e-v\rVert_2^2\leq\gamma$ will be inactive. For the case $\hat{y}=\argmin_{j} L_j(x,\theta)\neq y$, this constraint will be active (\ie, $\lVert e-v\rVert_2^2=\gamma$) and we will have
\begin{align}
\lambda^2 = \frac{1}{\gamma}\sum_j (L_j(x,\theta)-\nu_j+\mu)^2\, .
\label{eq:app_lambda_l2_cls}
\end{align}
For indices $j$ s.t. $v_j>0$, we will have the dual variables $\nu_j=0$. In practice, we would like to avoid setting a $\gamma$ so large that it assigns a zero weight to the annotated class (\ie, $v_y$ should be positive). In this case,
for indices $j\neq y$ s.t. $v_j=0$, we will have $L_j(x,\theta)-\nu_j=-\mu$. Since $\mu$ is the mean of all $\{\nu_j-L_j(x,\theta)\}_{j=1}^k$ and a subset of these are equal to $\mu$ (\ie, indices $Z=\{j: v_j=0\}$), $\mu$ should also be the mean of the complementary subset $\bar{Z}$, \ie,
\begin{align}
\mu=\frac{1}{|\bar{Z}|}\sum_{j\in \bar{Z}} (\nu_j-L_j(x,\theta))= -\frac{1}{|\bar{Z}|}\sum_{j\in \bar{Z}} L_j(x,\theta)\,.
\label{eq:app_mu_l2_cls}
\end{align}
For the non-zero indices in $\bar{Z}\setminus y$, we will have $L_j(x,\theta)<-\mu$ (as $v_j>0$). For $v_y>0$, we have $L_y(x,\theta)<\lambda-\mu$, 
and for $v_y=0$ we will have $L_y=\lambda+\nu_y-\mu\geq\lambda-\mu$. 

The solution can be obtained by sorting the losses $\{L_j(x,\theta)\}_{j=1}^k$ in ascending order and doing a search for the nonzero indices $\bar{Z}$ over the possible solution set $S=\{[m]\cup\tilde{y}\}_{1\leq m<\tilde{y}}$, where $[m]=\{1,\ldots,m\}$, and $\tilde{y}$ denotes the index where the loss $L_y(x,\theta)$ falls in this ranking of losses. The number of candidate solutions are $\tilde{y}$ and the correct solution can be identified by checking certain conditions.
To check for correctness of a candidate solution $\bar{Z}$, we need to compute $\mu$ using Eq.~\eqref{eq:app_mu_l2_cls}, compute the nonzero $\nu_j$ using Eq.~\eqref{eq:app_v_l2_cls}, compute $\lambda$ using Eq.~\eqref{eq:app_lambda_l2_cls}, and ensure the indices in $\bar{Z}$ have positive values for $v_j$. 
A first heuristic that often provided correct solutions in our experiments was to compute the mean $\tilde{\mu}$ of the losses $\{L_j(x,\theta):L_j(x,\theta)\leq L_y(x,\theta)\}$ and set the nonzero indices of $v$ to be $\{j:L_j(x,\theta)<\tilde{\mu}\}\cup \{y\}$.

\subsection{Class reweighting with $\ell_\infty$-distance}
\label{app:linf_dist}
We now take $D_2$ to be the squared $\ell_\infty$-distance  and consider the following problem:
\begin{align}
\min_v \sum_j v_{j} L_j(x,\theta),\quad \text{s.t. }  \lVert e - v\rVert_\infty \leq \gamma, \sum_j v_{j}=1, v_{j} \geq 0\, .
\label{prob:app_class_w_inf}
\end{align}
For this problem, if $\hat{y}:=\argmin_j L_j(x,\theta)=y$ then the solution will trivially be the one-hot vector, \ie, $v=e$. If $\hat{y}\neq y$, then $v_y=1-\gamma$, $v_{\hat{y}}=\gamma$, and for other indices $j$, $v_j=0$. 

\subsection{Class reweighting with Reverse \kl-divergence}
\label{app:revkl_div}
Taking $D_2$ to be the reverse \kl~divergence $\sum_k e_{k} \log \frac{e_{k}}{v_{k}} = -\log v_y$, we have,
\begin{align}
\min_v \sum_j v_{j} L_j(x,\theta),\quad \text{s.t. }  -\log v_y \leq \gamma, \sum_j v_{j}=1, v_{j} \geq 0\, .
\label{prob:app_class_w_ikl}
\end{align}
The first condition is satisfied for any distribution $v$ where $v_y \geq \exp(-\gamma)$. This problem as similar solution as in the case of total variation and $\ell_\infty$ distance. If $\hat{y}:=\argmin_j L_j(x,\theta)=y$ then the solution will trivially be the one-hot vector, \ie, $v=e$. If $\hat{y}\neq y$, then $v_y=1-\exp(-\gamma)$, $v_{\hat{y}}=\exp(-\gamma)$, and for other indices $j$, $v_j=0$.

\section{Experimental setup for Two Moons dataset}
\label{app:twomoons}
We consider $1000$ samples from the Two Moons dataset\footnote{\url{https://scikit-learn.org/stable/modules/generated/sklearn.datasets.make_moons.html}} with noise standard deviation of $0.05$. We corrupt $30\%$ of the labels by randomly flipping the class. We also normalize the input features by subtracting the mean and dividing by the standard deviation. We train a two layer fully connected neural network with $10$ and $20$ hidden units, respectively with $\tanh$ activation function. The output layer predicts the class probability using a sigmoid activation function and the loss is defined as the binary CE between the output probabilities and the $0/1$ labels. We train the network using a SGD optimizer with learning rate equal to $0.05$ and heavy ball momentum equal to $0.9$. We set the batch size to $10$
 and train the model for $20$ epochs.
 
We apply a burn-in period of $6$ epochs where we train the model with the baseline CE loss and without any reweighting. For CIW, we use apply the weighting using $\alpha$-divergence with $\alpha=0.5$ and $\mu=0.5$.
 
\section{Algorithmic details}
\label{app:algorithm}
We give an algorithmic sketch of the proposed methods CIW, CICW, and CICW-M in Algorithm \ref{alg:ciw}, Algorithm \ref{alg:cicw}, and Algorithm \ref{alg:cicw-m}, respectively. 

\begin{algorithm}[H]
\caption{Constrained Instance Reweighting (CIW)}\label{alg:ciw}
\begin{algorithmic}[1]
\State \textbf{Hyperparameters:} $\alpha$ for $\alpha$-divergence, $\lambda$ or $\mu$ in Eq. (5), (6), or (7) in the main paper, burn-in parameter $b$ 
\State \textbf{Init:} Model parameters $\theta=\theta^{(0)}$
\State \textbf{For} $t = 1 ~\text{to}~ T$: ~~~~// training iterations
\State ~~~~ Get a minibatch $\{(x_i,y_i)\}_{i=1}^n$
\State ~~~~ Compute base-loss (\eg, cross-entropy loss) $L(x_i,y_i,\theta^{(t-1)})$ $\forall\,i$
\State ~~~~ \textbf{If} $t>b$:
\State ~~~~~~~~ Compute weights $w_i$ using Eq. (5), (6), or (7) in the main paper, depending on the \mbox{value of $\alpha$}
\State ~~~~~~~~  $w_i = $ \texttt{stop-gradient}($w_i$) $\forall\,i$
\State ~~~~ \textbf{Else:}
\State ~~~~~~~~ $w_i = \frac{1}{n}$ $\forall\,i$
\State ~~~~ Compute reweighted loss $\sum_i w_i L(x_i,y_i,\theta^{(t-1)})$
\State ~~~~ Update model parameters to $\theta^{(t)}$ using gradient of the reweighted loss
\end{algorithmic}
\end{algorithm}

\begin{algorithm}[H]
\caption{Constrained Instance and Class Reweighting (CICW)}\label{alg:cicw}
\begin{algorithmic}[1]
\State \textbf{Hyperparameters:} $\alpha$ for $\alpha$-divergence, $\lambda$ or $\mu$ in Eq. (5), (6), or (7) in the main paper, burn-in parameter $b$, divergence $D_2$ and parameter $\gamma$ in Eq. (8) in the main paper. 
\State \textbf{Init:} Model parameters $\theta=\theta^{(0)}$
\State \textbf{For} $t = 1 ~\text{to}~ T$: ~~~~// training iterations
\State ~~~~ Get a minibatch $\{(x_i,y_i)\}_{i=1}^n$
\State ~~~~ Compute base-loss (\eg, cross-entropy loss) $L(x_i,y_i,\theta^{(t-1)})$ $\forall\,i$
\State ~~~~ \textbf{If} $t>b$:
\State ~~~~~~~~~ \multiline{Compute the class reweighted loss $\tilde{L}(x_i,\theta^{(t-1)})=\sum_j v_j L(x_i,j,\theta^{(t-1)})$ by solving Eq. (9) (for $D_2=\ell_1$) or Eq. (11) (for $D_2=\ell_2$)}
\State ~~~~~~~~~ \multiline{Compute weights $w_i$ using Eq. (5), (6), or (7) in the main paper (depending on the value of $\alpha$), with class reweighted losses $\tilde{L}(x_i,\theta^{(t-1)})$}
\State ~~~~~~~  $w_i = $ \texttt{stop-gradient}($w_i$) $\forall\,i$
\State ~~~~ \textbf{Else:}
\State ~~~~~~~~ $\tilde{L}(x_i,\theta^{(t-1)})=L(x_i,y_i,\theta^{(t-1)})$ $\forall\,i$
\State ~~~~~~~~ $w_i = \frac{1}{n}$ $\forall\,i$
\State ~~~~ Compute reweighted loss $\sum_i w_i \tilde{L}(x_i,\theta^{(t-1)})$
\State ~~~~ Update model parameters to $\theta^{(t)}$ using gradient of the reweighted loss
\end{algorithmic}
\end{algorithm}

\begin{algorithm}[H]
\caption{Constrained Instance and Class Reweighting with Mixup (CICW-M)}\label{alg:cicw-m}
\begin{algorithmic}[1]
\State \textbf{Hyperparameters:} $\alpha$ for $\alpha$-divergence, $\lambda$ or $\mu$ in Eq. (5), (6), or (7) in the main paper, burn-in parameter $b$, divergence $D_2$ and parameter $\gamma$ in Eq. (8) in the main paper, Mixup-type (IW-Mix/SIW-Mix) and Reweighting (Mixup-base/Mixup-reweight) in Sec. 2.4 in the main paper. 
\State \textbf{Init:} Model parameters $\theta=\theta^{(0)}$
\State \textbf{For} $t = 1 ~\text{to}~ T$: ~~~~// training iterations
\State ~~~~ Get a minibatch $\{(x_i,y_i)\}_{i=1}^n$
\State ~~~~ Compute base-loss (\eg, cross-entropy loss) $L(x_i,y_i,\theta^{(t-1)})$ $\forall\,i$
\State ~~~~ \textbf{If} $t>b$:
\State ~~~~~~~~~ \multiline{Compute the class reweighted loss $\tilde{L}(x_i,\theta^{(t-1)})=\sum_j v_j L(x_i,j,\theta^{(t-1)})$ by solving Eq. (9) (for $D_2=\ell_1$) or Eq. (11) (for $D_2=\ell_2$)}
\State ~~~~~~~~~ \multiline{Compute weights $w_i$ using Eq. (5), (6), or (7) in the main paper (depending on the value of $\alpha$), with class reweighted losses $\tilde{L}(x_i,\theta^{(t-1)})$}
\State ~~~~~~~  $w_i = $ \texttt{stop-gradient}($w_i$) $\forall\,i$
\State ~~~~~~~~~ \multiline{Generate mixed-up minibatch ($X^{(m)}, Y^{(m)}$) using weights $w_i$ depending on the Mixup-type, as described in Sec. 2.4}
\State ~~~~~~~~~ \multiline{Compute base-loss for mixed-up examples, \ie, $L(X^{(m)}_i,\theta^{(t-1)}):=L(X^{(m)}_i,Y^{(m)}_i,\theta^{(t-1)})$ $\forall\,i$}
\State ~~~~~~~ \textbf{If} Mixup-base:
\State ~~~~~~~~~~ $\tilde{L}_i(\theta^{(t-1)})=L(X^{(m)}_i,\theta^{(t-1)})$
\State ~~~~~~~~~~ $w_i = \frac{1}{n}$ $\forall\,i$
\State ~~~~~~~ \textbf{If} Mixup-reweight:
\State ~~~~~~~~~~~~~~ \multiline{Compute class reweighted loss $\tilde{L}(X^{(m)}_i,\theta^{(t-1)})$ and instance weights $w_i$ for mixed-up examples using Lines 7--9 of Algorithm \ref{alg:cicw}}
\State ~~~~~~~~~~~ $\tilde{L}_i(\theta^{(t-1)})=\tilde{L}(X^{(m)}_i,\theta^{(t-1)})$
\State ~~~~ \textbf{Else:}
\State ~~~~~~~~ $\tilde{L}_i(\theta^{(t-1)})=L(x_i,y_i,\theta^{(t-1)})$ $\forall\,i$
\State ~~~~~~~~ $w_i = \frac{1}{n}$ $\forall\,i$
\State ~~~~ Compute reweighted loss $\sum_i w_i \tilde{L}_i(\theta^{(t-1)})$
\State ~~~~ Update model parameters to $\theta^{(t)}$ using gradient of the reweighted loss
\end{algorithmic}
\end{algorithm}

\section{Optimization}
\label{app:opt_details}

We train the ResNet-18 model for both CIFAR datasets using SGD optimizer with a momentum of $0.9$ with Nesterov acceleration. We use initial learning rate of 0.1 and a piecewise constant learning rate schedule of $(10^{-2}, 10^{-3}, 10^{-4})$ at $(30,80,110)$ epochs, respectively. We use a batch size of $128$ and train the model for $140$ epochs. For the ResNet50v2 model for Clothing1M, we again use a SGD optimizer with a momentum of $0.9$ with Nesterov acceleration and additionally use a weight decay. We tune the initial learning rate and weight decay parameter for the cross-entropy baseline using Clothing1M validation set and fix them to these values for all other methods ($0.005$ for learning rate and $10^{-5}$ for weight decay). We use a batch size of 64 and finetune it for total 8 epochs, while  reducing the learning rate to $0.0005$ after 5 epochs. %

\section{Symmetric and Asymmetric label noise}
\label{app:label_noise}

To simulate symmetric noise, we follow earlier works and flip the label of each training and validation example to the other labels with probability $\eta$.
However, this differs from some earlier works such as \cite{arazo2019unsupervised} where the noisy label is \emph{randomly sampled} from all possible labels which results in effectively a lower noise rate as there is $1/k$ chance of sampling the correct label for $k$ classes.

To generate asymmetric noise, we follow the procedure in~\citep{ma2020normalized}. For CIFAR-10, we consider the following mapping between the classes: `truck' $\rightarrow$ `automobile', `bird' $\rightarrow$ `airplane', `deer' $\rightarrow$ `horse', `cat' $\leftrightarrow$ `dog'. For CIFAR-10, only the subset of classes that conform to the noise pattern is noisified, thus naturally dividing the noise rate by half. For CIFAR-100, the super-classes are adopted from the original description of the dataset at~\url{https://www.cs.toronto.edu/~kriz/cifar.html} and the classes within in each super-class are mapped to each other with a certain probability.

\section{Experiments: other $\alpha$-divergences}
\label{app:exp_alpha}
We report the complete results, including results for other $\alpha$-divergences for $\alpha\in\{0,0.5,1\}$, in Table \ref{app:tab:cifar_sym}. All $\alpha$-divergences show similar empirical performance when the hyperparameter $\lambda$ is tuned on the noisy validation set, with \kl-divergence $(\alpha=1)$ performing slightly better or on-par with others. 

\begin{table*}[!ht]
\vspace{-4mm}
\caption{{\small \bf Test accuracy  on CIFAR-10 and CIFAR-100 with symmetric label noise using ResNet-18.} Noise rate is varied in the range $\{0.2, 0.4, 0.6, 0.8\}$. The results (mean$\pm$std) are reported over five random runs and the top two results are highlighted in {\bf boldface}.
Proposed methods are CIW, CICW and CICW-M. 
Blank cells for CICW and CICW-M denote that we do not get improvement over CIW with a $\gamma>0$ (Obj. \eqref{prob:class_w}).
Methods marked $^{\boldsymbol{\ddagger}}$ need to either maintain scalar/vector weights over full train set, or fit a model on loss values of full train set (extra forward pass for entire train set per epoch).}
\label{app:tab:cifar_sym}
\centering
\small
\resizebox{.85\textwidth}{!}{
\begin{tabular}{c|ccccc}
\toprule
 \multirow{2}{*}{Methods} &  \multicolumn{5}{c}{Noise Rate ({\bf CIFAR-10})}\\
&  Clean & 0.2 & 0.4 & 0.6 & 0.8 \\ \hline \hline
  CE & 92.44 $\pm$ 0.17 & 84.03 $\pm$ 0.38 & 78.56 $\pm$ 0.98 & 67.64 $\pm$ 1.12 & 36.86 $\pm$ 1.96 \\
  Bi-tempered \citep{bitempered}& 92.55 $\pm$ 0.10 & 89.74 $\pm$ 0.22 & 83.87 $\pm$ 0.37 & 72.49 $\pm$ 0.77 & 40.41 $\pm$ 1.09 \\
  APNL \citep{ma2020normalized} & 90.93 $\pm$ 0.16 & 88.63 $\pm$ 0.35 & 84.97 $\pm$ 0.35 & 77.47 $\pm$ 0.46 & 42.71 $\pm$ 1.47 \\ \midrule
   EG$^{\boldsymbol{\ddagger}}$ \citep{bar2021multiplicative,majidi2021exponentiated} & 92.29 $\pm$ 0.12 & 90.22 $\pm$ 0.16 & 86.48 $\pm$ 0.33 & 80.03 $\pm$ 0.74 & 45.38 $\pm$ 2.24 \\
   ELR$^{\boldsymbol{\ddagger}}$ \citep{liu2020early} & -- & 91.15 $\pm$ 0.05 & 88.38 $\pm$ 0.44& 78.94 $\pm$ 0.47& 40.15 $\pm$ 3.28\\
 \midrule
 CIW ($\alpha=0$) & 92.10 $\pm$ 0.16 & 88.18 $\pm$ 0.16 & 84.32 $\pm$ 0.27 & 75.16 $\pm$ 0.60 & 43.12 $\pm$ 0.64 \\
 CIW ($\alpha=0.5$) & 91.66 $\pm$ 0.21 & 88.67 $\pm$ 0.21 & 85.64 $\pm$ 0.24 & 77.84 $\pm$ 0.28 & 42.65 $\pm$ 1.16 \\
 CIW ($\alpha=1$) & 92.14 $\pm$ 0.20& 88.74 $\pm$ 0.26 & 85.13 $\pm$ 0.28 & 78.27 $\pm$ 0.41 & 43.07 $\pm$ 0.34 \\
\midrule
 CICW ($\alpha=1$) & -- & 89.49 $\pm$ 0.14 & 86.45 $\pm$ 0.34 & 78.68 $\pm$ 0.37 & 44.03 $\pm$ 1.87 \\
 \midrule
Mixup \citep{zhang2017mixup} & 93.66 $\pm$ 0.20 & 89.48 $\pm$ 0.24 & 85.00 $\pm$ 0.46 & 75.12 $\pm$ 0.67 & 44.44 $\pm$ 1.96 \\
Dyn-Mixup$^{\boldsymbol{\ddagger}}$ \citep{arazo2019unsupervised} & 92.52 $\pm$ 0.22 & 89.69 $\pm$ 0.15 & 87.30 $\pm$ 0.33 & 78.04 $\pm$ 0.59 & 46.04 $\pm$ 0.88 \\ \midrule 
CICW-M ($\alpha=1$) & -- & {\bf 91.34 $\pm$ 0.16} & {\bf 89.95 $\pm$ 0.15} & {\bf 84.53 $\pm$ 0.27} & {\bf 58.74 $\pm$ 0.73} \\
\midrule
\midrule
&  \multicolumn{5}{c}{Noise Rate ({\bf CIFAR-100})}\\
&  Clean & 0.2 & 0.4 & 0.6 & 0.8 \\ \hline \hline
  CE & 70.52 $\pm$ 0.32 & 56.09 $\pm$ 0.39 & 43.34 $\pm$ 1.17 & 29.41 $\pm$ 1.00 & 12.47 $\pm$ 1.09 \\
  Bi-tempered \citep{bitempered} & 72.01 $\pm$ 0.26 & 67.51 $\pm$ 0.29 & 60.60 $\pm$ 0.71 & 47.25 $\pm$ 0.74 & 20.89 $\pm$ 0.64 \\
  APNL \citep{ma2020normalized}& 67.62 $\pm$  0.45 & 65.66 $\pm$ 0.37 & 60.25 $\pm$ 0.43 & 47.28 $\pm$ 0.97 & 20.12 $\pm$ 0.93\\ \midrule
   EG$^{\boldsymbol{\ddagger}}$ \citep{bar2021multiplicative,majidi2021exponentiated} & 70.38 $\pm$ 0.19 & 67.15 $\pm$ 0.35 & 61.59 $\pm$ 0.32 & 50.58 $\pm$ 0.58 & 25.76 $\pm$ 0.61 \\
   ELR$^{\boldsymbol{\ddagger}}$ \citep{liu2020early} & -- & 65.73 $\pm$ 0.24& 57.37 $\pm$ 0.48& 40.64 $\pm$0.51& 12.87 $\pm$ 0.57 \\
 \midrule
 CIW ($\alpha=0$) & 69.66 $\pm$ 0.44 & 64.30 $\pm$ 0.21 & 57.60 $\pm$ 0.45 & 45.00 $\pm$ 0.94 & 23.01 $\pm$ 0.58 \\
 CIW ($\alpha=0.5$) & 68.08 $\pm$ 0.51& 64.47 $\pm$ 0.40 & 58.95 $\pm$ 0.17 & 49.71 $\pm$ 0.39 & 25.36 $\pm$ 0.19 \\
 CIW ($\alpha=1$) & 70.08 $\pm$ 0.44 & 65.89 $\pm$ 0.30        & 60.54 $\pm$ 0.45 & 50.53 $\pm$ 0.37 & 25.00 $\pm$ 0.80 \\
\midrule
 CICW ($\alpha=1$) & -- & -- & 60.85 $\pm$ 0.44 & 50.57 $\pm$ 0.46 & 25.36 $\pm$ 0.92 \\
 \midrule
Mixup \citep{zhang2017mixup} & 72.63 $\pm$ 0.22 & 65.48 $\pm$ 0.27 & 58.61 $\pm$ 0.32  & 45.46 $\pm$ 0.94 & 21.86 $\pm$ 0.65 \\
Dyn-Mixup$^{\boldsymbol{\ddagger}}$ \citep{arazo2019unsupervised} & 71.28 $\pm$ 0.33 & 67.74 $\pm$ 0.24 & 61.14 $\pm$ 0.48 & 49.37 $\pm$ 0.48 & 26.48 $\pm$ 1.19 \\ \midrule 
CICW-M ($\alpha=1$) & -- & {\bf 68.23 $\pm$ 0.23} & {\bf 63.64 $\pm$ 0.45} & {\bf 54.46 $\pm$ 0.72} & {\bf 28.94 $\pm$ 0.74} \\
\bottomrule
\end{tabular}
}
\vspace{-0.3cm}
\end{table*}

\section{Hyperparameter selection}
\label{app:hyperparams}
As described in the main paper, we do the hyperparameter selection using the noisy validation sets. 
Our assumption of availability of a noisy validation set is realistic and is often encountered in practice.
We perform hyperparameter search for all methods using the validation set.  

Our methods have following hyperparameters: (i) CIW has one hyperparameter, $\lambda$ for \kl-divergence, (Eq.~\eqref{eq:w_kldiv}), or hyperparameter $\mu$ for reverse-\kl~divergence (Eq.~\eqref{eq:w_revkl}) and $\alpha$-divergence for $\alpha=0.5$ (Eq.~\eqref{eq:w_alphadiv_raw}). (ii) CICW has two additional hyperparameters: the choice of divergence $D_2$ (total-variation or $\ell_2$), and parameter $\gamma$ in the objective \eqref{prob:class_w}, however in our experiments we did not see a major difference in performance with total-variation vs $\ell_2$ distance, so it is fine to fix $D_2$ and only tune for $\gamma$. In Tables \ref{tab:cifar10_hyp}, \ref{tab:cifar100_hyp}, \ref{tab:cifar10_hyp_asym} and \ref{tab:cifar100_hyp_asym} we still report the $D_2$ and $\gamma$ that are automatically selected using the noisy validation set. (iii) CICW-M has two additional hyperparameters over CICW: the choice of how to use importance weights in Mixup (IW-Mix or SIW-Mix in Sec.~\ref{subsec:mixup}), and the choice of how to use mixed up examples for training (Mixup-base or Mixup-reweight in Sec.~\ref{subsec:mixup}). However, in the case of symmetric noise, we fix $\alpha=1$, and fix $D_2$ to \kl-divergence for CIFAR-10 and  %
fix it to total variation distance for CIFAR-100. 
In this case of asymmetric noise, we use Dynamic-CICW-M which has three hyperparameters -- $\alpha$ and $\lambda$ for CIW, and  $\beta$ for mixup. However, we fix $\alpha=1$ in this case, leaving us with two hyperparameters. 

We use a \emph{burn-in} period of 4000 minibatch iterations for CIFAR-10 and CIFAR-100 where we train the network using standard cross-entropy loss before activating the instance and class reweighting. We use a burn-in period of 2000 iterations for Clothing1M. 

We provide the selected hyperparameter for completeness, which were obtained using the noisy validation set (obtained after splitting the noisy train set into 90\%/10\% for train and validation, respectively). Hyperparameters and their description for various methods are provided in Tables \ref{tab:cifar10_hyp} and \ref{tab:cifar100_hyp} for the case of symmetric noise and in Tables \ref{tab:cifar10_hyp_asym} and \ref{tab:cifar100_hyp_asym} for the case of asymmetric noise. 
For \emph{Dyn-Mixup}, $\alpha=\beta$ parameters are used only for mixing labels, and the images are mixed using the posterior probabilities (divided by the sum) from the $\beta$-mixture fitted on the training losses.\footnote{\url{https://github.com/PaulAlbert31/LabelNoiseCorrection/blob/master/train.py\#L175}
}
\begin{table*}[!ht]
\caption{{\bf Hyperparameter settings for CIFAR-10 with symmetric label noise.~} \emph{Bi-tempered:} hyperparameters are ($t_1$, $t_2$, $m$ final iteration when annealing $t_1$ and $t_2$ from $1$ to the selected temperatures). \emph{APNL:} hyperparameters are (active loss type, passive loss type, weight on the active loss, $\gamma$ parameter for focal-loss). \emph{EG:}  hyperparameters are ($\eta$ learning rate, $0 \leq \lambda \leq 1$ regularizer, $m$ final epoch when annealing learning rate from $0$). \emph{CIW:} single hyperparameter $\lambda$ or $\mu$ in Eq.~(5), (6), (7) of the main paper. \emph{CICW:} hyperparameters are ($\lambda$ or $\mu$, $D_2$, $\gamma$), refer to Eq.~(8) in the main paper. \emph{Mixup:} ($\alpha=\beta$ for beta-distribution). \emph{Dyn-Mixup:} ($\alpha=\beta$ for beta-distribution). \emph{CICW-M:} hyperparameters are ($\lambda$ or $\mu$, $\gamma$, Mixup-type, Reweight), Mixup-type is one of IW-Mix or SIW-Mix, and Reweight is Yes (Mixup-reweight) or No (Mixup-base). Refer to Sec. 2.4 in the main paper for Mixup details.}
\label{tab:cifar10_hyp}
\centering
\small
\resizebox{.95\textwidth}{!}{
\begin{tabular}{c|cccc}
\toprule
 \multirow{2}{*}{Methods} &  \multicolumn{4}{c}{Noise Rate }\\
& 0.2 & 0.4 & 0.6 & 0.8 \\ \hline \hline
  Bi-tempered \citep{bitempered} (1-$t_1$, $t_2$, $m$) &  (0.001, 6.3, 20k)  & (0.3, 5.0, 20k) & (0.9, 10.0, 20k)& (0.5, 10.0, 20k)\\
  APNL \citep{ma2020normalized} (AL, PL, $w_a$, $\gamma_{\text{FL}}$)& (nfl, mae, .8, 3.)& (nfl, mae, 0.7, 5.) & (nfl, mae, 0.7, 5.) & (nfl, mae, 0.7, 8.) \\ \midrule
   EG \citep{bar2021multiplicative,majidi2021exponentiated} ($\eta$, 1-$\lambda$, $m$) & (0.05, 0.0, 5) & (0.05, 0.01, 5) & (0.15, 0.05, 10) & (0.1, 0.0, 10)\\
 \midrule
 CIW, $\alpha=1$, ($\lambda$) & 0.2 & 0.1 & 1.5 & 1.5 \\
\midrule
 CICW, $\alpha=1$ ($\lambda, D_2, \gamma$)  & (0.8, $\ell_1$, 0.01) & (1, $\ell_2$, 0.1) & (1, $\ell_2$, 0.12) & (1.3, $\ell_1$, 0.05) \\
 \midrule
Mixup \citep{zhang2017mixup} ($\alpha=\beta$) & 1 & 5 & 5 & 2 \\
Dyn-Mixup \citep{arazo2019unsupervised} ($\alpha=\beta$) & 2 &  30 & 30 & 30 \\ \midrule 
\multirow{2}{*}{\parbox{4cm}{\centering CICW-M, $\alpha=1, D_2=\text{KL}$, ($\lambda, \gamma$, Mixup-type, Reweight)}} & \multirow{2}{*}{(3, 0.01, IW, Y)} & \multirow{2}{*}{(2.5, 0.01, SIW, Y)}& \multirow{2}{*}{(2.5, 0.1, SIW, Y)}& \multirow{2}{*}{(2.5, 0.1, SIW, Y)} \\ \\
\bottomrule
\end{tabular}
}
\end{table*}

\begin{table*}[h]
\caption{{\bf Hyperparameter settings for CIFAR-100 with symmetric label noise.~} \emph{Bi-tempered:} hyperparameters are ($t_1$, $t_2$, $m$ final iteration when annealing $t_1$ and $t_2$ from $1$ to the selected temperatures). \emph{APNL:} hyperparameters are (active loss type, passive loss type, weight on the active loss, $\gamma$ parameter for focal-loss). \emph{EG:}  hyperparameters are ($\eta$ learning rate, $0 \leq \lambda \leq 1$ regularizer, $m$ final epoch when annealing learning rate from $0$). \emph{CIW:} single hyperparameter $\lambda$ or $\mu$ in Eq.~(5), (6), (7) of the main paper. \emph{CICW:} hyperparameters are ($\lambda$ or $\mu$, $D_2$, $\gamma$), refer to Eq.~(8) in the main paper. \emph{Mixup:} ($\alpha=\beta$ for beta-distribution). \emph{Dyn-Mixup:} ($\alpha=\beta$ for beta-distribution). \emph{CICW-M:} hyperparameters are ($\lambda$ or $\mu$, $\gamma$, Mixup-type, Reweight), Mixup-type is one of IW-Mix or SIW-Mix, and Reweight is Yes (Mixup-reweight) or No (Mixup-base). Refer to Sec. 2.4 in the main paper for Mixup details.}

\label{tab:cifar100_hyp}
\centering
\small
\resizebox{.95\textwidth}{!}{
\begin{tabular}{c|cccc}
\toprule
 \multirow{2}{*}{Methods} &  \multicolumn{4}{c}{Noise Rate }\\
& 0.2 & 0.4 & 0.6 & 0.8 \\ \hline \hline
  Bi-tempered \citep{bitempered} (1-$t_1$, $t_2$, $m$) &  (0.01, 3.0, 20k)  & (0.1, 3.0, 20k) & (0.6, 3.0, 20k) & (1.0, 2.0, 20k)\\
  APNL \citep{ma2020normalized} (AL, PL, $w_a$, $\gamma_{\text{FL}}$)& ((nfl, mae, 0.99, 8.0))& (nfl, mae, 0.99, 10) & (nfl, mae, 0.98, 5) & (nfl, mae, 0.95, 8) \\ \midrule
   EG \citep{bar2021multiplicative,majidi2021exponentiated} ($\eta$, 1-$\lambda$, $m$) & (0.05, 0.01, 10) & (0.1, 0.05, 10) & (0.1, 0.05, 10) & (0.1, 0.05, 10)\\
 \midrule
 CIW, $\alpha=1$, ($\lambda$) & 0.75 & 1.0 & 1.5 & 1.5 \\
\midrule
 CICW, $\alpha=1$ ($\lambda, D_2, \gamma$)  & (0.75, --, 0) & (1, $\ell_1$, 0.1) & (1.5, $\ell_1$, 0.1) & (1.2, $\ell_2$, 0.1) \\
 \midrule
Mixup \citep{zhang2017mixup} ($\alpha=\beta$) & 2 & 2 & 1 & 1 \\
Dyn-Mixup \citep{arazo2019unsupervised} ($\alpha=\beta$) & 2 & 5 & 2 & 5 \\ \midrule 
\multirow{2}{*}{\parbox{4cm}{\centering CICW-M, $\alpha=1$, $D_2=\ell_1$ ($\lambda, \gamma$, Mixup-type, Reweight)}} & \multirow{2}{*}{(0.9, 0, IW, N)} & \multirow{2}{*}{(2, 0.05, SIW, N)}& \multirow{2}{*}{(1.7, 0.02, SIW, N)}& \multirow{2}{*}{(1.8, 0.02, SIW, N)} \\ \\
\bottomrule
\end{tabular}
}
\end{table*}

\begin{table*}[!ht]
\caption{{\bf Hyperparameter settings for CIFAR-10 with asymmetric label noise.~} \emph{Bi-tempered:} hyperparameters are ($t_1$, $t_2$, $m$ final iteration when annealing $t_1$ and $t_2$ from $1$ to the selected temperatures). \emph{APNL:} hyperparameters are (active loss type, passive loss type, weight on the active loss, $\gamma$ parameter for focal-loss). \emph{EG:}  hyperparameters are ($\eta$ learning rate, $0 \leq \lambda \leq 1$ regularizer, $m$ final epoch when annealing learning rate from $0$). \emph{CIW:} single hyperparameter $\lambda$ or $\mu$ in Eq.~(5), (6), (7) of the main paper. \emph{CICW:} hyperparameters are ($\lambda$ or $\mu$, $D_2$, $\gamma$), refer to Eq.~(8) in the main paper. \emph{ELR:} hyperparameters are (decay factor $\beta$, $\lambda_{\text{ELR}}$ regularizer). \emph{CIW-ELR:} hyperparameters are ($\beta$, $\lambda_{\text{ELR}}$, CIW $\alpha$, $\lambda$ or $\mu$). \emph{Mixup:} ($\alpha=\beta$ for beta-distribution). \emph{Dyn-Mixup:} ($\alpha=\beta$ for beta-distribution). \emph{Dyn-CICW-M:} we fix CIW $\alpha=1$ and the remaining hyperparameters are ($\lambda$ or $\mu$, Mixup $\beta$).}
\label{tab:cifar10_hyp_asym}
\centering
\small
\resizebox{.95\textwidth}{!}{
\begin{tabular}{c|cccc}
\toprule
 \multirow{2}{*}{Methods} &  \multicolumn{4}{c}{Noise Rate }\\
& 0.1 & 0.2 & 0.3 & 0.4 \\ \hline \hline
  Bi-tempered \citep{bitempered} (1-$t_1$, $t_2$, $m$) &  (0.02, 10.0, 50k)  & (0.01, 10.0, 50k) & (0.01, 20.0, 50k) & (0.01, 15.0, 50k)\\
  APNL \citep{ma2020normalized} (AL, PL, $w_a$, $\gamma_{\text{FL}}$)& (nfl, rce, 0.8, 1.0) & (nfl, mae, 0.7, 2.0) & (nfl, mae, 0.6, 0.5) & (nce, mae, 0.7, --) \\ \midrule
   EG \citep{bar2021multiplicative,majidi2021exponentiated} ($\eta$, 1-$\lambda$, $m$) & (0.1, 0.05, 10)  & (0.1, 0.01, 10) & (0.15, 0.02, 10) & (0.2, 0.02, 10)\\
 \midrule
 CIW, $\alpha=1$, ($\lambda$) & 1.0 & 0.8 & 0.1 & 0.1\\
\midrule
 CICW, $\alpha=1$ ($\lambda, D_2, \gamma$)  & (0.5, KL, 0.1) & (0.5, KL, 0.1) & (0.1, --, 0.0) & (0.1, KL, 0.02)\\
 \midrule
 ELR \citep{liu2020early} ($\beta$, $\lambda_{\text{ELR}}$)&  (0.9, 5.0) & (0.9, 5.0) & (0.8, 3.5) & (0.9, 6.0)\\
 CIW-ELR ($\beta$, $\lambda_{\text{ELR}}$, $\alpha$, $\lambda$) & (0.9, 4.0, 1, 10.0) & (0.9, 3.0, 0, 2.0) & (0.6, 2.0, -2, 0.4) & (0.7, 1.5, -2, 0.02)\\
 \midrule
Mixup \citep{zhang2017mixup} ($\alpha=\beta$) & 2 & 5 & 5 & 5\\
Dyn-Mixup \citep{arazo2019unsupervised} ($\alpha=\beta$) & 1 & 2 & 2 & 1\\ \midrule 
Dyn-CICW-M, $\alpha=1$ ($\lambda$, Mixup $\beta$) & (20.0, 0.5) & (20.0, 1.0) & (12.0, 0.5) & (10.0, 0.2)\\
\bottomrule
\end{tabular}
}
\end{table*}

\begin{table*}[h]
\caption{{\bf Hyperparameter settings for CIFAR-100 with asymmetric label noise.~} \emph{Bi-tempered:} hyperparameters are ($t_1$, $t_2$, $m$ final iteration when annealing $t_1$ and $t_2$ from $1$ to the selected temperatures). \emph{APNL:} hyperparameters are (active loss type, passive loss type, weight on the active loss, $\gamma$ parameter for focal-loss). \emph{EG:}  hyperparameters are ($\eta$ learning rate, $0 \leq \lambda \leq 1$ regularizer, $m$ final epoch when annealing learning rate from $0$). \emph{CIW:} single hyperparameter $\lambda$ or $\mu$ in Eq.~(5), (6), (7) of the main paper. \emph{CICW:} hyperparameters are ($\lambda$ or $\mu$, $D_2$, $\gamma$), refer to Eq.~(8) in the main paper. \emph{ELR:} hyperparameters are (decay factor $\beta$, $\lambda_{\text{ELR}}$ regularizer). \emph{CIW-ELR:} hyperparameters are ($\beta$, $\lambda_{\text{ELR}}$, CIW $\alpha$, $\lambda$ or $\mu$). \emph{Mixup:} ($\alpha=\beta$ for beta-distribution). \emph{Dyn-Mixup:} ($\alpha=\beta$ for beta-distribution). \emph{Dyn-CICW-M:} we fix CIW $\alpha=1$ and the remaining hyperparameters are ($\lambda$, Mixup $\beta$).}
\label{tab:cifar100_hyp_asym}
\centering
\small
\resizebox{.95\textwidth}{!}{
\begin{tabular}{c|cccc}
\toprule
 \multirow{2}{*}{Methods} &  \multicolumn{4}{c}{Noise Rate }\\
& 0.1 & 0.2 & 0.3 & 0.4 \\ \hline \hline
  Bi-tempered \citep{bitempered} (1-$t_1$, $t_2$, $m$) & (0.5, 5.0, 50k) & (0.001, 5.0, 50k) & (0.02, 5.0, 50k) & (0.001, 5.0, 50k)\\
  APNL \citep{ma2020normalized} (AL, PL, $w_a$, $\gamma_{\text{FL}}$) & (nfl, mae, 0.99, 1.0) & (nfl, rce, 0.99, 8.0) & (nfl, mae, 0.99, 1.0) &  (nfl, mae, 0.99, 1.0)\\ \midrule
   EG \citep{bar2021multiplicative,majidi2021exponentiated} ($\eta$, 1-$\lambda$, $m$)  & (0.05, 0.01, 10) & (0.1, 0.0, 10) & (0.1, 0.02, 10) & (0.15, 0.05, 10)\\
 \midrule
 CIW, $\alpha=1$, ($\lambda$)  & 0.2 & 0.1 & 0.1 & 0.1\\
\midrule
 CICW, $\alpha=1$ ($\lambda, D_2, \gamma$)   & (0.1, $\ell_1$, 0.2) & (0.1, $\ell_1$, 0.2) &  (0.1, $\ell_1$, 0.05) & (0.1, $\ell_1$, 0.3)\\
 \midrule
 ELR \citep{liu2020early} ($\beta$, $\lambda_{\text{ELR}}$)&  (0.9, 8.0) & (0.9, 10.0) & (0.9, 10.0) & (0.9, 10.0)\\
 CIW-ELR ($\beta$, $\lambda_{\text{ELR}}$, $\alpha$, $\lambda$) & (0.9, 2.0, 1, 1.5) & (0.8, 4.0, 0, 1.0) & (0.8, 3.0, 0, 1.0) & (0.8, 4.0, 0, 1.0)\\
 \midrule
Mixup \citep{zhang2017mixup} ($\alpha=\beta$)  & 2.0 & 2.0 & 5.0 & 5.0\\
Dyn-Mixup \citep{arazo2019unsupervised} ($\alpha=\beta$)  & 2.0 & 1.0 & 2.0 & 2.0\\  \midrule 
Dyn-CICW-M, $\alpha=1$, ($\lambda$, Mixup $\beta$) & (5.0, 0.5) & (20.0, 1.0) & (20.0, 1.0) & (20.0, 1.0)\\
\bottomrule
\end{tabular}
}
\end{table*}

\subsection{Hyperparameter sensitivity}
\label{app:hyper_sensitivity}
\begin{figure*}[t!]
\begin{center}
 \subfigure[\normalsize CIW sensitivity to $\lambda$]{\includegraphics[width=0.45\textwidth]{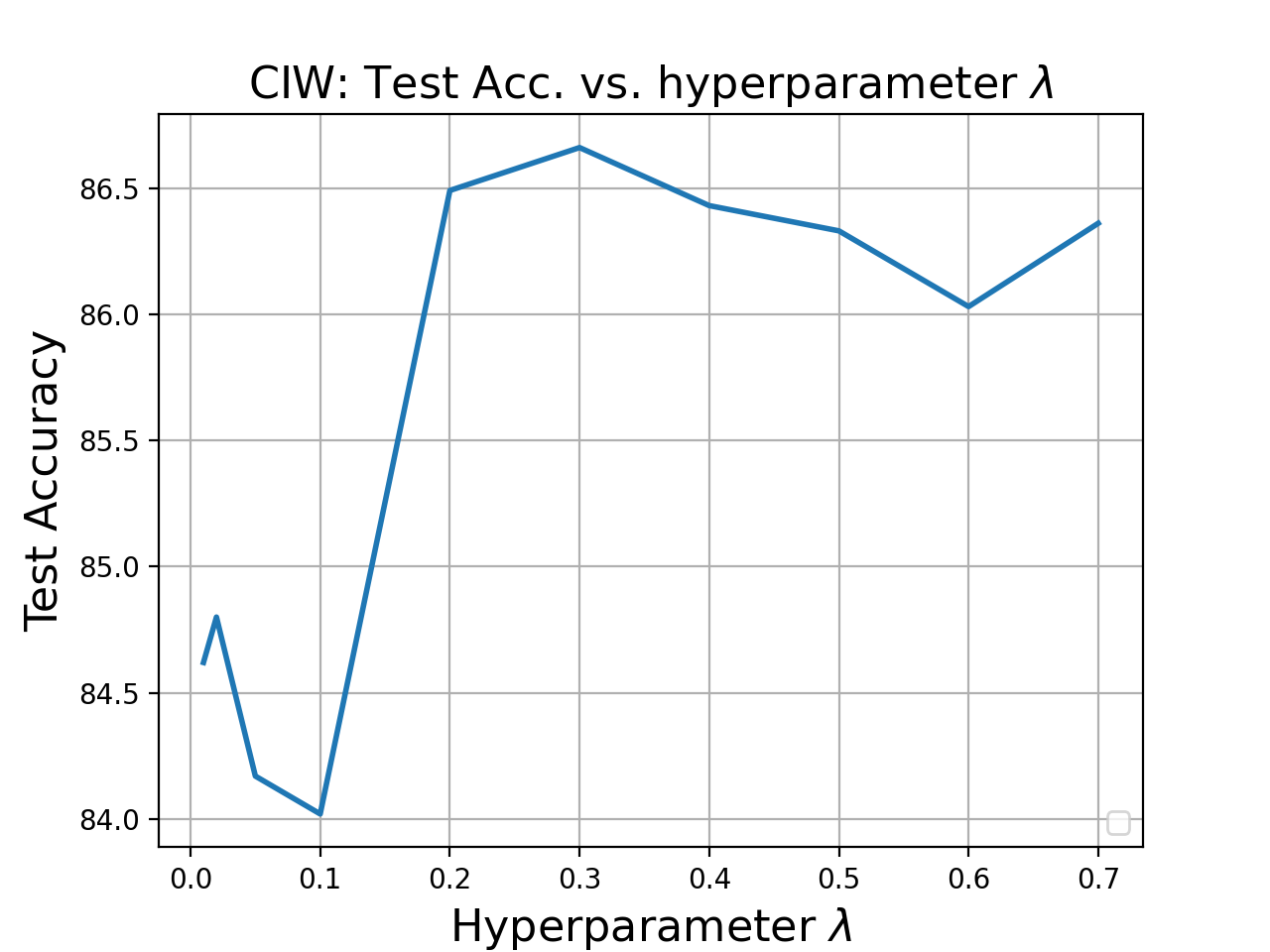}}\hspace{1mm}
 \subfigure[\normalsize CICW sensitivity to $\gamma$, for fixed $\lambda=1$]{\includegraphics[width=0.45\textwidth]{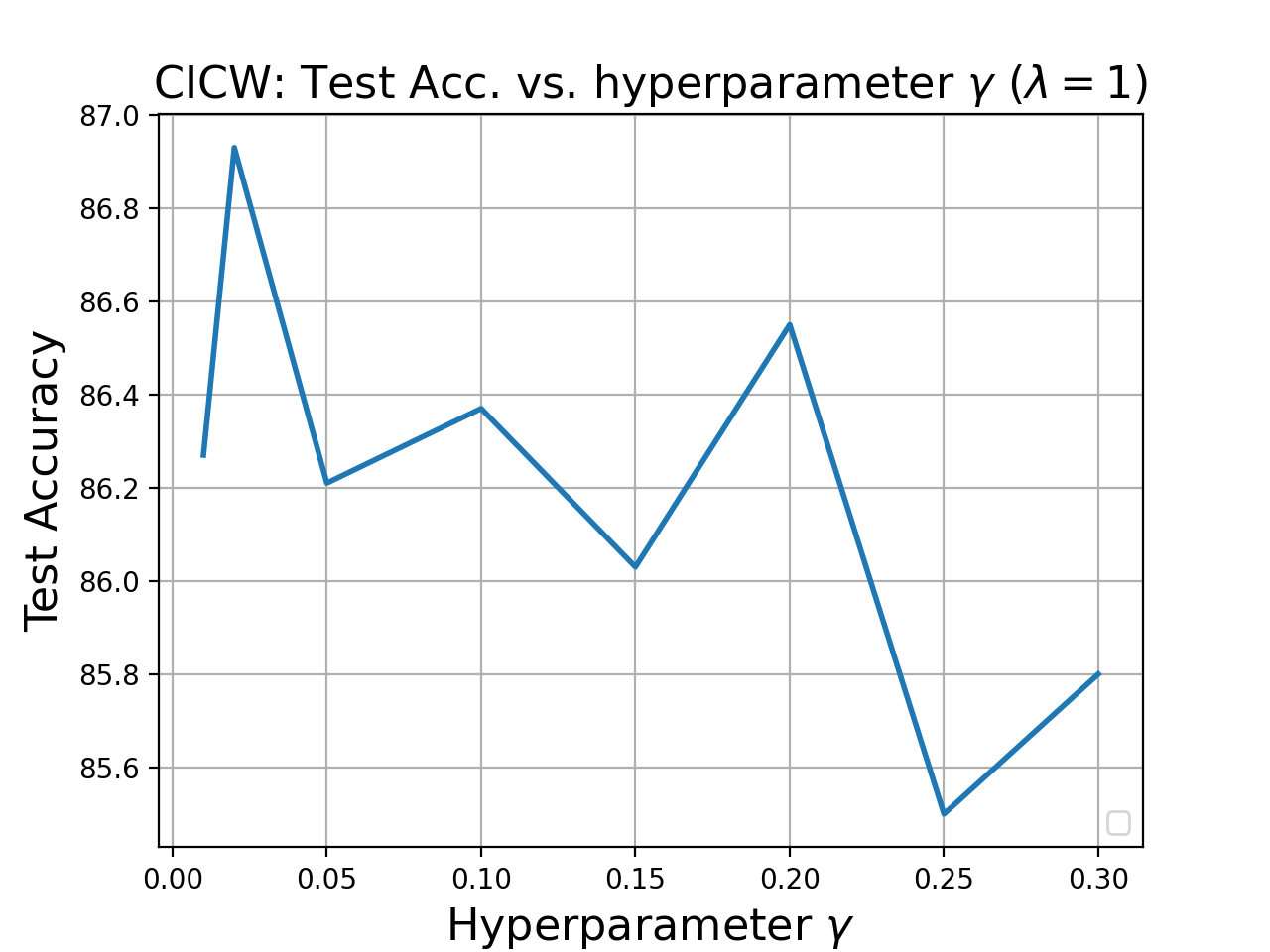}}
\caption{Variation in test accuracy as a function of hyperparameters for CIFAR-10 and symmetric noise rate $\eta=0.4$: (a) Sensitivity of CIW to changes in hyperparameter $\lambda$, (b) Sensitivity of CICW to changes in hyperparameter $\gamma$ for fixed $\lambda=1$. Both CIW and CICW are reasonably robust to hyperparameter variation (within $\sim 2\%$).}
\label{app:fig:hyper_sensitivity}
\end{center}
\end{figure*}

We observe that both CICW and CICW-M are reasonably robust to hyperparameter variability. We show the changes in test accuracy as a function of hyperparameters in Figure \ref{app:fig:hyper_sensitivity} for both CIW and CICW. Both CIW and CICW are reasonably robust to hyperparameter variation with test accuracy varying within $\sim 2.5\%$ range.

\end{document}